\documentclass[journal]{IEEEtran}
\usepackage{amsmath,amsfonts}
\usepackage{bbding}
\usepackage{verbatim}
\usepackage{graphicx}
\usepackage{float}
\usepackage{url}
\usepackage{booktabs, subfig}
\usepackage[table]{xcolor}
\usepackage{threeparttable}
\usepackage{makecell}
\usepackage{multirow}
\usepackage{subfig}
\usepackage[numbers,sort&compress]{natbib}
\definecolor{lightgreen}{RGB}{169,209,142}
\definecolor{lightblue}{RGB}{155,192,226}
\definecolor{lightorange}{RGB}{244,177,131}
\definecolor{lightyellow}{RGB}{255,217,102}
\usepackage[colorlinks,
            linkcolor=red,
            citecolor=blue
            ]{hyperref}

\def\SPSB#1#2{\rlap{\textsuperscript{\textcolor{black}{#1}}}\SB{#2}}
\def\SB#1{\textsubscript{\textcolor{black}{#1}}}

\begin{document}

\title{Adaptive Loose Optimization for Robust Question Answering}

\author{
    Jie Ma,~\IEEEmembership{Member,~IEEE,}
    Pinghui Wang, ~\IEEEmembership{Senior Member,~IEEE,}
    Zewei Wang\textsuperscript{$\dagger$},
    Dechen Kong\textsuperscript{$\dagger$},
    Min Hu,
    Ting Han,
    and
    Jun Liu, ~\IEEEmembership{Senior Member,~IEEE},
\thanks{ $\dagger$ denotes the authors contribute equally to the work.} 
\thanks{ Jie Ma and Pinghui Wang are with the  Ministry of Education of Key Laboratory for Intelligent Networks and Network Security, School of Cyber Science and Engineering, Xi’an Jiaotong University, Xi’an, Shaanxi 710049, China.}
\thanks{ Ting Han, Zewei Wang, and Dechen Kong are with the  Ministry of Education of Key Laboratory for Intelligent Networks and Network Security, School of Automation Science and Engineering, Xi'an Jiaotong University, Xi'an, Shaanxi 710049, China.}
\thanks{Min Hu is with the China Mobile Research Institute, China.}
\thanks{ Jun Liu is with the Shannxi Provincial Key Laboratory of Big Data Knowledge Engineering, School of Computer Science and Technology, Xi'an Jiaotong University, Xi'an, Shaanxi 710049, China.}
\thanks{This work was supported by the National Key Research and Development Program of China (2021YFB1715600), the National Natural Science Foundation of China (U22B2019, 62272372, 62293553, 62250066, 621737002).}
}

\markboth{Journal of \LaTeX\ Class Files,~Vol.~14, No.~8, August~2021}%
{Shell \MakeLowercase{\textit{et al.}}: A Sample Article Using IEEEtran.cls for IEEE Journals}



\maketitle

\begin{abstract}
  Question answering methods are well-known for leveraging data bias, such as the language prior in visual question answering and the position bias in machine reading comprehension (extractive question answering). Current debiasing methods often come at the cost of significant in-distribution performance to achieve favorable out-of-distribution generalization ability, while non-debiasing methods sacrifice a considerable amount of out-of-distribution performance in order to obtain high in-distribution performance. Therefore, it is challenging for them to deal with the complicated changing real-world situations. In this paper, we propose a simple yet effective novel loss function with adaptive loose optimization, which seeks to make the best of both worlds for question answering. Our main technical contribution is to reduce the loss adaptively according to the ratio between the previous and the current optimization state on mini-batch training data. This loose optimization can be used to prevent non-debiasing methods from overlearning data bias while enabling debiasing methods to maintain slight bias learning. Experiments on the visual question answering datasets, including VQA v2, VQA-CP v1, VQA-CP v2, GQA-OOD, and the extractive question answering dataset SQuAD demonstrate that our approach enables QA methods to obtain state-of-the-art in- and out-of-distribution performance in most cases. The source code has been released publicly in \url{https://github.com/reml-group/ALO}.
\end{abstract}

\begin{IEEEkeywords}
Visual question answering, extractive question answering, language and position bias, multi-modality learning, debiasing, robustness.
\end{IEEEkeywords}

\section{Introduction}
\label{intro}
\IEEEPARstart{Q}{uestion} Answering (QA) requires machines to answer questions accurately given a specific context. In general, the context in extractive QA \cite{rajpurkar2016squad,hu2019read} is natural language, while in visual QA \cite{antol2015vqa,yanvqa,9525040} it is an image. In recent years, there has been a notable surge of interest and research into QA techniques, resulting in their widespread implementation throughout various fields. These applications encompass a wide range of domains, including but not limited to customer service chatbots, personal assistants, and search engines \cite{zhang2023toward}. 

It is frequently observed that current QA methods prefer to excessively exploit the training bias or statistical regularity, which bypasses the context comprehension for a short-cut answer \cite{rubi,niu2021counterfactual}. For instance, in order to achieve high accuracy for questions involving the color of roses, visual QA methods may not learn the proper behavior when the majority of the roses are red. It is much simpler to learn from the bias connecting the words \emph{what}, \emph{color}, and \emph{roses} with the most frequent answer \emph{red}, as opposed to carefully looking at the image, identifying a rose (\includegraphics[height=0.6\baselineskip]{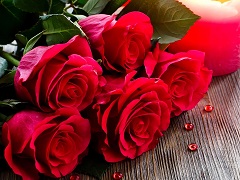}, \includegraphics[height=0.6\baselineskip]{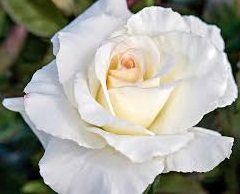}), and determining its color (\emph{red}, \emph{white}). This phenomenon is demonstrated in \cite{antol2015vqa}, wherein their model, which does not utilize the image context, achieved an overall accuracy of 48.09\%. Likewise, extractive QA methods may use the bias of positional cues to locate the answer in passages \cite{ko2020look}. \emph{Therefore, QA methods that exhibit strong performance in in-distribution situations ineluctably fail in out-of-distribution test scenarios.} The former refers to cases where the distribution of a dataset is similar to that observed during training, whereas the latter pertains to distributions that differ significantly or even oppose those encountered at training time.
\begin{figure*}[tbp]
    \centering  
    \includegraphics[width=0.7\columnwidth]{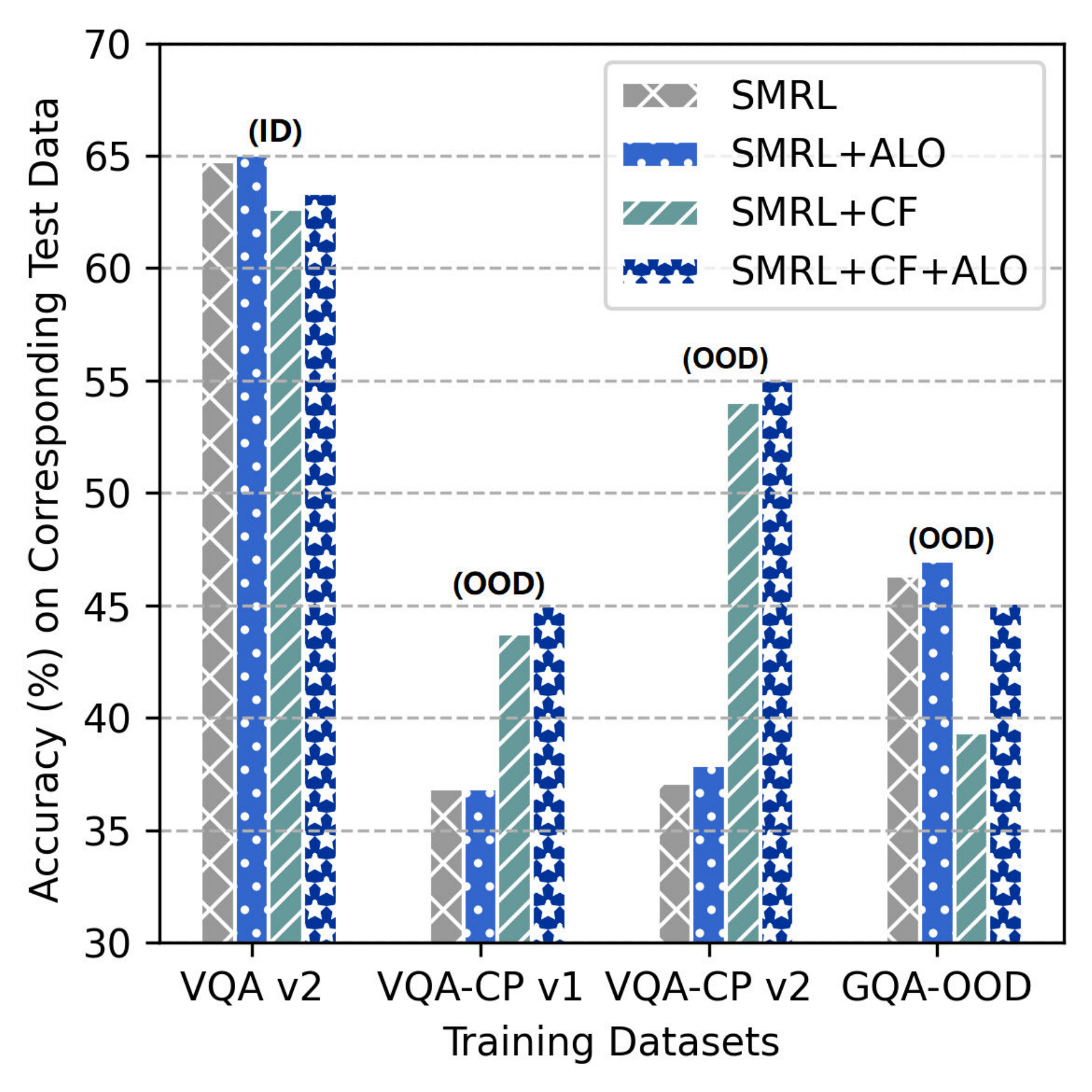}
    \hspace{1.5cm}
    \includegraphics[width=0.7\columnwidth]{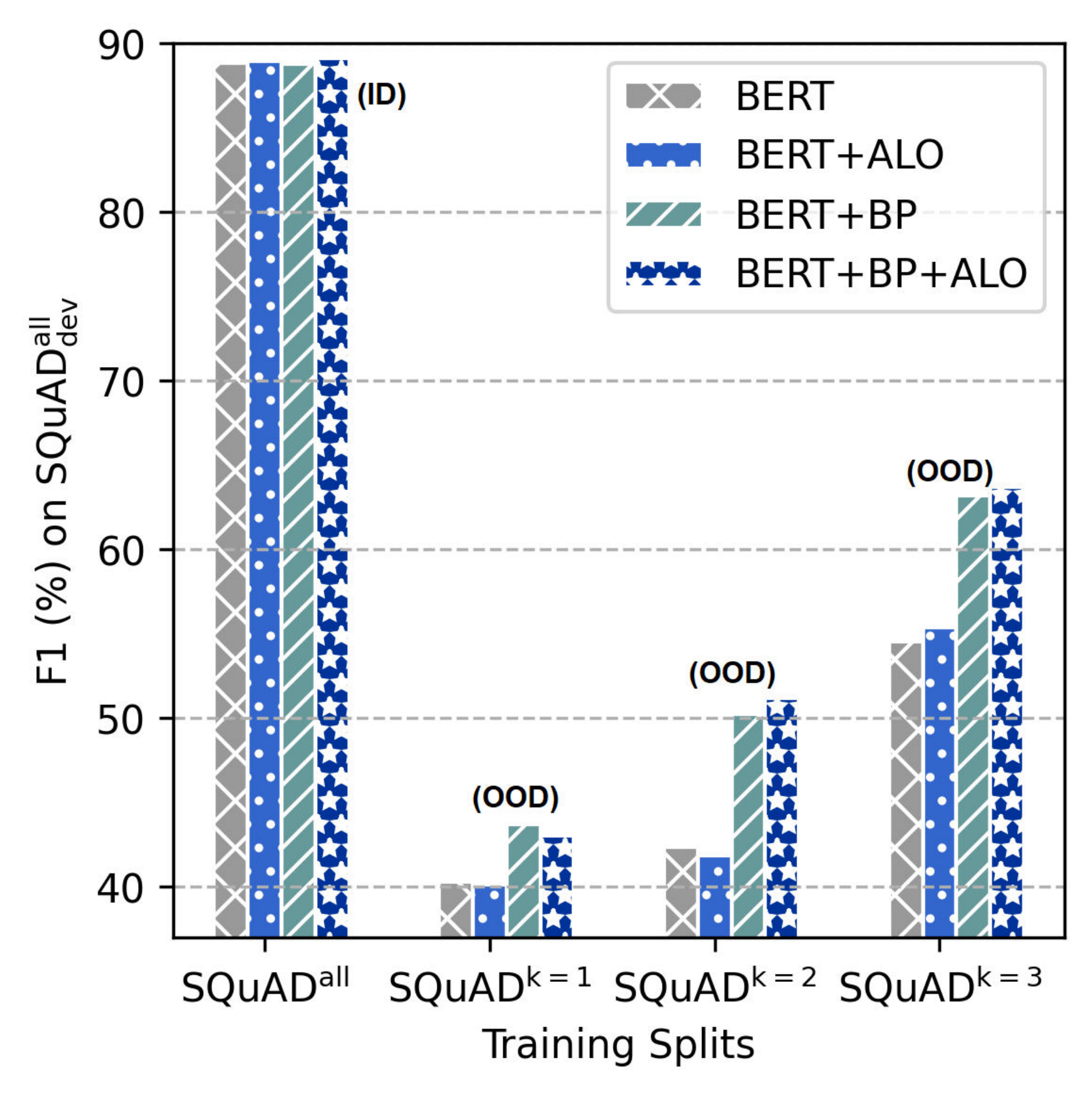}
    \caption{Performance comparison of visual and extractive QA methods on various situations. Current debiasing QA methods such as CF variant \protect\cite{niu2021counterfactual} and BERT+BP \protect\cite{hinton2002training} obtain high out-of-distribution but low in-distribution performance. By contrast, the non-debiasing methods such as SMRL \protect\cite{cadene2019murel} and BERT display impressive in-distribution performance but struggle with out-of-distribution situations. Our proposed Adaptive Loose Optimization (ALO), allows non-debiasing and debiasing methods to obtain improvements in both in- and out-of-distribution situations. The left (right) shows the results of visual QA (extractive QA). All other datasets are out-of-distribution scenarios, with the exception of VQA v2 \protect\cite{goyal2017making} and SQuAD \protect\cite{rajpurkar2016squad}.}
	\label{fig:vqa-qa}
\end{figure*}

Recently, several QA debiasing solutions including data augmentation \cite{goyal2017making}, self-supervised learning \cite{selvaraju2019}, and ensemble learning \cite{shah2019cycle,ClarkYZ19} were proposed to mitigate the bias (language or position) learning, and achieved success on out-of-distribution situations. However, the assumption that the training and test distributions are drastically different, or even the opposite, causes them to sacrifice the in-distribution performance. Therefore, we can observe a very intriguing phenomenon shown in Fig. \ref{fig:vqa-qa}: debiasing methods perform well in out-of-distribution but fail in in-distribution scenarios, while non-debiasing methods achieve high in-distribution but low out-of-distribution performance. \emph{This observation prompts us to consider whether we can achieve the best of both worlds.}

Consider why these methods work well in only one of the in-distribution and out-of-distribution scenarios, from an optimization perspective. One possible explanation could be that all methods pursue their own distinct objectives through excessively rigorous optimization. In general, they regard visual and extractive QA as multiclass classification problems and use a standard entropy loss function to achieve their respective goals. Specifically, non-debiasing methods are allowed to connect words such as \emph{what}, \emph{color}, and \emph{roses} with the answer \emph{red}, whereas debiasing methods are not. However, excessively rigorous optimization would hurt their capacity to handle a variety of scenarios. Taking non-debiasing visual QA methods as an example, a loss of $-\log0.3 \approx 0.52$ will still be used to optimize them if the probability for answer \emph{red} is 0.3 and the prediction is correct over the large answer space (usually 3,129 candidate answers within VQA v2.0). This will result in overly strong connections (bias), such as those between critical words of questions and answers, leading to failure in the out-of-distribution situation. Similarly, the debiasing methods optimized by a rigorous loss function excessively weaken the connection, causing failure in the in-distribution scenario. Therefore, the mentioned connection (bias) should not be excessively explored or weakened for the non-debiasing and debiasing methods, respectively.

Motivated by this, we propose a novel loss function with Adaptive Loose Optimization (ALO), which serves as a more effective alternative to previous QA approaches for handling in- and out-of-distribution scenarios. This loss function reduces the vanilla probability $\hat{y}$ dynamically based on the loose factor $\gamma \in (0,1)$.
In other words, the loss function transforms $\hat{y}$ to $\hat{y}^{\gamma}$, which enables non-debiasing methods to decrease bias learning and debiasing methods to maintain slight bias learning. It is apparent that a smaller controlling factor results in looser optimization. To derive the factor and achieve adaptive loose optimization, we leverage the ratio of the previous and current optimization states on mini-batch training data. Notably, the adaptive loose optimization does not harm the original capacity of the methods, such as the debiasing ability of debiasing methods, but rather drives them to perform well in  both in- and out-of-distribution situations simultaneously. Experimental results show that the combinations of previous QA methods and adaptive loose optimization achieve better in-distribution (visual QA: VQA v2 \cite{goyal2017making}; extractive QA: SQuAD \cite{rajpurkar2016squad}.) and out-of-distribution (visual QA: VQA-CP v1, VQA-CP v2 \cite{agrawal2018don}, and GQA-OOD \cite{kervadec2021roses}; extractive QA:  SQuAD with changing priors \cite{anderson2018bottom}) performance in most cases. 

The contributions of this paper can be summarized as follows. 
\begin{itemize}
	\item We introduce a novel loss function with adaptive loose optimization that empowers QA methods to achieve robust performance.
	\item We conduct extensive and thorough experiments on five publicly available datasets in the fields of visual and extractive QA. Additionally, we also perform theoretical analyses to demonstrate the effectiveness of our proposed approach.
	\item Our empirical study reveals that combining non-debiasing (debiasing) QA methods with adaptive loose optimization may yield strong performance in both in- and out-of-distribution scenarios.
\end{itemize}

The remainder of this paper is organized as follows. Section \ref{sec:rel} introduces the related works about robust visual QA, robust extractive QA, and robustness estimation. The combination of visual and extractive QA methods with the adaptive loose optimization is described respectively in Section \ref{sec:met}. Section \ref{sec:exp} presents the empirical study on visual and extractive QA datasets under in- and out-of-distribution settings and shows the theoretical analyses of our approach from a gradient perspective. The concluding remarks are shown in Section \ref{sec:con}.

\section{Related Work}
\label{sec:rel}
\subsection{Visual QA}
Previous methods \cite{cadene2019murel, anderson2018bottom, 9525040, jiang2020defense, garderes2020conceptbert, guo2021re} have made significant progress on in-distribution datasets such as VQA v1 and VQA v2. The methods are classified into four groups \cite{wu2017visual}: joint embedding-based, attention mechanism-based, compositional, and external knowledge enhanced. Joint embedding-based methods \cite{noh2016image,gao2015you} usually leverage recurrent and convolutional neural networks to learn the representations of questions and images in a common space. This enables one to perform answer prediction by feeding them into a classifier. However, the mentioned methods may bring noise into the prediction stage. To address the issue, attention mechanism-based methods \cite{9466370,chen2022grounding} fuse their representations at the object and word level. Compositional methods \cite{andreas2016learning} allow for customized computations for each instance of a question, while external knowledge-enhanced methods aim to enhance question and image understanding by querying knowledge bases and incorporating relevant prior or professional knowledge.

Recent studies \cite{rubi, ClarkYZ19, chen2020counterfactual} found that bias in training data is the major driving force behind the visual QA methods. For example, \emph{tennis} is the correct answer for 41\% of the questions starting with \emph{what sport is}, and \emph{2} is the correct answer for 39\% of the questions starting with \emph{how many} in VQA v1 \cite{antol2015vqa}. To address this issue, numerous debiasing approaches are proposed and assessed on out-of-distribution datasets. The datasets include VQA-CP v1, VQA-CP v2, and GQA-OOD, whose distributions between training and test splits are significantly different or even reversed. Current debiasing methods can be divided into three groups: ensemble learning, self-supervised learning, and data augmentation.

Ensemble learning-based methods utilize a debiasing branch along with a visual QA model to predict answers comprehensively. To address the issue of language bias learning, Clark et al. \cite{ClarkYZ19} employed a pipeline strategy in which they first trained a model that only utilized questions as input to learn bias, and then used the bias-only model to train an ensemble. By contrast, RUBi \cite{rubi} uses a question-only branch to capture language bias, affecting visual QA methods to prevent the bias learned from questions in an end-to-end manner. Inspired by causal effect, Niu et al. \cite{niu2021counterfactual} unified RUBi into a counterfactual learning framework, which defines language bias as the direct causal effect of questions on answers, and mitigates the bias by subtracting the direct language effect from total causal effect. Similarly, Li et al. \cite{li2022invariant} proposed an invariant grounding framework that enables the differentiation of causal scenes and the highlighting of their causal impact on the answer. The analysis presented in \cite{niu2021counterfactual, niu2021introspective, Kolling} indicates that ensemble learning-based methods achieved superior performance compared to other approaches. Based on the findings, we aim to explore the potential benefits of integrating them with our proposed method.

Data augmentation-based methods \cite{chen2022rethinking} leverage data annotating or generating to reduce the bias. Teney et al.\cite{NEURIPS20200} partitioned the data into well-chosen, non-independent-and-identically-distributed subsets and trained visual QA backbones on different environments to improve the out-of-distribution generalization ability. To generate numerous counterfactual training samples, Chen et al. \cite{chen2020counterfactual} proposed a model-agnostic counterfactual sample synthesizing scheme, which generates samples by masking critical objects in images or words in questions. Liang et al. \cite{liang2020learning} employed the mentioned scheme to generate counterfactual samples and introduced a contrastive learning mechanism to improve the generalization ability. However, previous methods have primarily focused on exploring the robustness of visual QA from the perspective of analyzing language context. Gupta et al. \cite{gupta2022swapmix} took a different approach and examined the robustness of visual QA from the perspective of visual context. They proposed a straightforward perturbation strategy that involves swapping object features with those of another object. This forces the model to concentrate more on the relevant objects and less on irrelevant contextual information, ultimately improving the overall robustness of the system. The above-generated samples often contain errors and appear unnatural. In response, Chen et al. \cite{chen2022rethinking} proposed a knowledge distillation-based data augmentation method, which can generate pseudo answers for all composed image-question pairs.

Self-supervised learning-based methods aim at enhancing the grounding ability by mining potential annotations. Specifically, Shah et al. \cite{shah2019cycle} proposed a model-agnostic training technique to incorporate cycle consistency in VQA methods, thus making them robust to linguistic variances and self-aware of their shortcomings. Si et al. \cite{si2022towards} proposed a robust visual QA system, which firstly distinguishes the biased and unbiased samples, and then learns a more general multimodal representation based on contrastive learning. However, the above-mentioned methods only explore bias elimination in the forward chaining ignoring the backward chaining. Motivated by this issue, Lao et al. \cite{lao2022vqa} proposed a bidirectional chaining approach. In their approach, backward chaining is employed to generate crucial visual facts selected in the forward chaining process, with the answer serving as the driving force. By incorporating hard-negative contrastive learning, this method can help alleviate excessive bias learning and improve overall performance. Although the debiasing methods discussed above result in a sustainable improvement in out-of-distribution performance, they come with the disadvantage of compromised in-distribution performance. 

\subsection{Extractive QA}
Previous methods \cite{xiong2018dcn, yu2018qanet, chen2021bidirectional, li2021asynchronous,3052594} regard the extractive QA as predicting the start and end positions of answers and have achieved sustainable success. These methods usually consist of four modules: embedding, feature extraction, context-question interaction, and answer prediction. Based on the direction of context-question interaction, the methods can be categorized into two groups: unidirectional and bidirectional. 

The unidirectional interaction methods aim to model the relationship from the query to the context, thereby emphasizing the most pertinent segments of the context based on the question. In particular, Wang et al. \cite{wang2017machine} made use of match-LSTM and pointer networks \cite{vinyals2015pointer} to perform span prediction over the context representation. Wang et al. \cite{wang2017gated} proposed a gated self-matching network, which employs a self-matching attention mechanism to refine the context representation. Nonetheless, unidirectional interaction methods disregard the question words that are crucial for answer prediction. To tackle this limitation, a plethora of bidirectional interaction approaches have been proposed. For instance, Seo et al. \cite{SeoKFH17} developed a bidirectional attention flow network, which utilizes the context-to-query and query-to-context attention mechanisms to model the relationship between questions and contexts. However, this method does not account for which parts of the context and question are focused on. To address this issue, Hu et al. \cite{hu2018reinforced} proposed a re-attention mechanism to refine past attention. In recent years, the advent of Transformer has spurred the development of numerous Transformer-based methods, including large language models, to tackle the extractive QA task. These methods leverage multi-head attention mechanisms, allowing them to model the bidirectional relationship between questions and contexts effectively. For instance, Zhao et al. \cite{yuqanet} combined local convolutions with global self-attention to accelerate answer prediction. Chen et al. \cite{chen2021adaptive} proposed an adaptive bidirectional attention mechanism to exploit multi-granularity sequence representations. Additionally, large language models like BERT and GPT have been proposed to handle multiple natural language processing tasks simultaneously, and have achieved significant success.
\begin{figure*}[tbp]
	\centering  
	\includegraphics[scale=0.53]{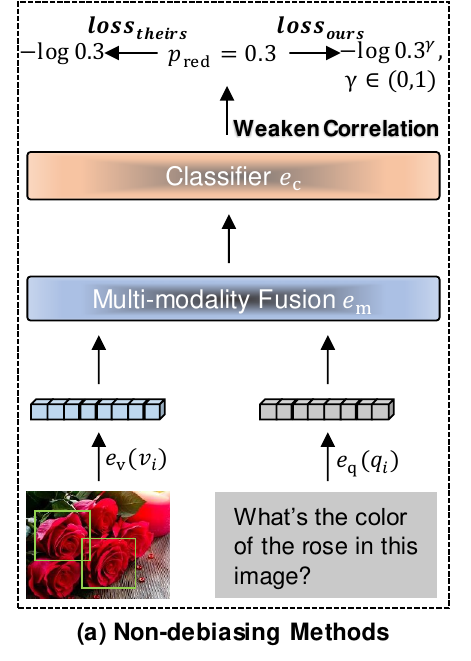}
	\hspace{0.5cm}
	\includegraphics[scale=0.53]{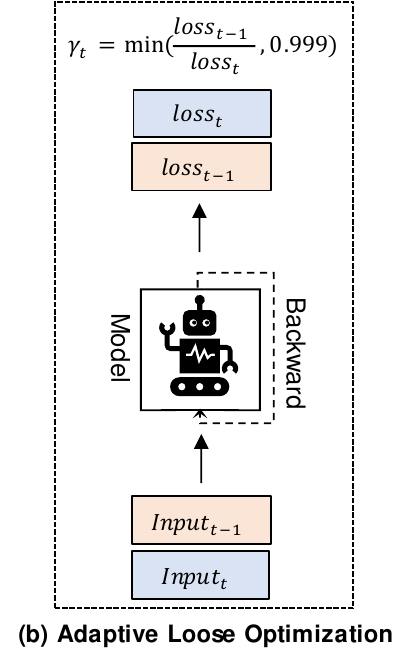}
	\hspace{0.5cm}
	\includegraphics[scale=0.53]{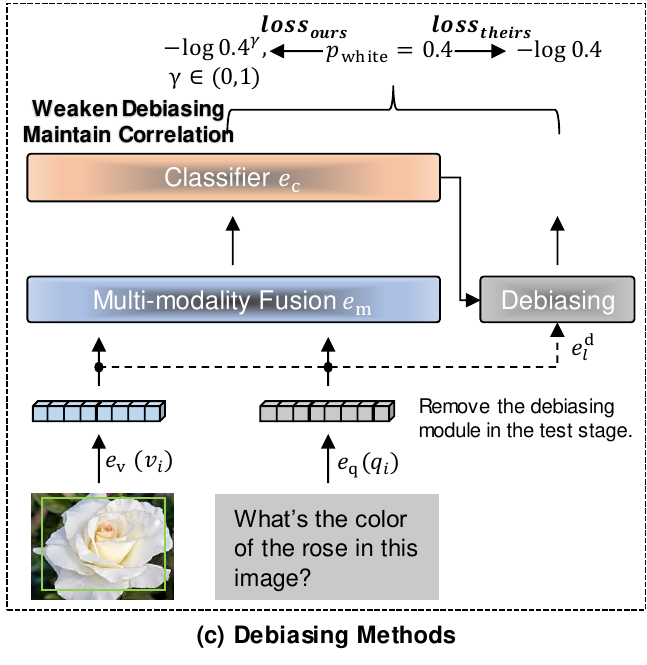}
	\caption{Comparison of non-debiasing and debiasing visual QA methods with or without adaptive loose optimization. The loose degree is controlled dynamically by the ratio of the last $t-1$ and current  $t$ optimization states.}
	\label{fig:vqa-arc-comp}
\end{figure*}

However, recent studies \cite{ko2020look, shinoda2022look} have found that position bias has a significantly negative impact on prediction performance. For instance, BERT only achieves a 37.48\% F1 score on the SQuAD development split after being fine-tuned on a biased split where each answer only appears in the first sentence. To explore this problem and investigate how methods can be robust to position bias, Ko et al. \cite{ko2020look} first developed a variant of SQuAD that contains various biased training settings, and then verified the combination of Bias Product (BP) and Learned-Mixin (LM) \cite{ClarkYZ19} with vanilla extractive QA methods respectively on the variant. Based on this study, Shinoda et al. \cite{shinoda2022look} focused on constructing the biased learning model from the answer-prior and position-only perspectives respectively. Nevertheless, extractive QA experiences identical issues concerning non-debiasing and debiasing approaches as seen in visual QA methods. Contrastingly, our adaptive loose optimization can augment non-debiasing and debiasing QA methods, simultaneously enhancing their in- and out-of-distribution performance. 

\subsection{Robustness Estimation}
The development of robust loss functions, from inlier (easy examples) or outlier (hard examples) perspectives, has attracted a lot of attention. Concretely, Huber loss \cite{shinoda2022look} lowers the impact of outliers by down-weighing the loss of examples with large errors. Contrary to it, the focal loss function \cite{HastieTF09} addresses the class imbalance by down-weighting inliers using double manually adjusted hyper-parameters including a weighting factor $\alpha$ and a focusing factor $\gamma$. Guo et al. \cite{lin2017focal} introduced a fixed distinctive weight for each answer, which prevents harsh parameter updates due to hard mistakes while guiding easy mistakes toward influential model learning. In comparison, by down-weighting inliers and outliers simultaneously, our adaptive loose optimization allows non-biasing (debiasing) methods to weaken (maintain) slight bias learning dynamically without any manual adjustments.

\section{Methods}
\label{sec:met}
The comparison results in \cite{niu2021counterfactual, niu2021introspective, Kolling} indicate that ensemble-based debiasing methods significantly improve out-of-distribution performance in visual and extractive QA tasks. Consequently, our study focuses on exploring the combination of these debiasing methods with our proposed adaptive loose optimization. Furthermore, we also explore the effectiveness of non-debiasing methods in combination with our approach.

\subsection{Visual QA}
\subsubsection{Task Formulation} 
Given a dataset $\mathcal{S}_{\mathrm{\iota}}=\{(v_i, q_i, a_i)\}_{i=1}^{n}$ including $n$ triplets with an image $v_i \in \mathcal{V}_{\mathrm{\iota}}$, a natural language question $q_i \in \mathcal{Q}_{\mathrm{\iota}}$, and an answer $a_i \in \mathcal{A}_{\mathrm{\iota}}$, visual QA requires machines to optimize parameters $\theta_{\iota}$ of the function $f_{\mathrm{\iota}}(\mathcal{V}_{\mathrm{\iota}}, \mathcal{Q}_{\mathrm{\iota}};\theta_{\iota}) \rightarrow \mathbb{R}^{|\mathcal{A}_{\mathrm{\iota}}|}$ to predict answers accurately, where $|\mathcal{A}_{\mathrm{\iota}}|$ denotes the total answers in the dataset. In other words, we regard the visual QA task as a multi-class classification problem.

\subsubsection{Non-debiasing Methods} 
At present, non-debiasing lightweight (small) models for visual QA often utilize the two-stream architecture, which leverages different encoders to process image and question inputs separately. These methods \cite{anderson2018bottom,cadene2019murel} usually apply an encoder $e_{\mathrm{v}}: v_i \rightarrow \mathbb{R}^{n_{\mathrm{v}} \times d_{\mathrm{v}}}$ like Faster R-CNN to learn $n_{\mathrm{v}}$ region-level representations with $d_{\mathrm{v}}$ dimensions of images, an encoder $e_{\mathrm{q}}: q_i \rightarrow \mathbb{R}^{n_{\mathrm{q}} \times d_{\mathrm{q}}}$ such as LSTM to obtain $n_{\mathrm{q}}$ word-level representations with $d_{\mathrm{q}}$ dimensions of questions, a multi-modality encoder $e_{\mathrm{m}}: \mathbb{R}^{n_{\mathrm{v}} \times d_{\mathrm{v}}} \times \mathbb{R}^{n_{\mathrm{q}} \times d_{\mathrm{q}}} \rightarrow \mathbb{R}^{d_{\mathrm{m}}}$ such as question-guided attention mechanisms to output fusion representations with $d_{\mathrm{m}}$ dimensions, and a classifier $e_{\mathrm{c}}: \mathbb{R}^{d_{\mathrm{m}}} \rightarrow \mathbb{R}^{|\mathcal{A}_{\mathrm{\iota}}|}$ to obtain predictions over candidate answer sets. The vanilla architecture is shown in Fig. \ref{fig:vqa-arc-comp} (a) and can be formulated as follows:
\begin{align}
	f_{\mathrm{\iota}} \left(v_i, q_i; \theta_{\iota} \right) = e_{\mathrm{c}} \left(e_{\mathrm{m}} \left( e_{\mathrm{v}} \left(v_i \right), e_{\mathrm{q}} \left(q_i \right) \right) \right).
\end{align}

To optimize the parameters $\theta_{\iota}$ of $f_{\mathrm{\iota}}$, non-debiasing methods minimize a standard cross-entropy loss function $\mathrm{CE}_{\mathrm{\iota}}$:
\begin{align}
	\mathrm{CE}_{\mathrm{\iota}} = - \left[ a_i \right] \log\left( \mathrm{softmax} \left( f_{\mathrm{\iota}} \left(v_i, q_i \right) \right) \right),
\end{align}
where $[a_i]$ denotes the class index for the ground-truth answer $a_i$. The non-debiasing methods are prone to learning bias or statistical regularity from the training data \cite{agrawal2018don}, resulting in their high in-distribution but low out-of-distribution performance.

\subsubsection{Debiasing Methods} Ensemble-based debiasing methods \cite{rubi,niu2021counterfactual} for visual QA usually apply a debiasing strategy $e_{\mathrm{\iota}}^{\mathrm{d}}$ such as the question-only branch in \cite{rubi} to reduce the impact of bias over the prediction logit obtained from the non-debiasing method. To optimize the parameters $\theta^{\mathrm{d}}_{\mathrm{\iota}}$ of $f^{\mathrm{d}}_{\mathrm{\iota}}$, the debiasing methods also minimize a standard cross-entropy loss function $\mathrm{CE}^{\mathrm{d}}_{\mathrm{\iota}}$. The vanilla architecture is shown in Fig. \ref{fig:vqa-arc-comp} (c) and can be formulated as follows:
\begin{align}
	& f^{\mathrm{d}}_{\mathrm{\iota}}(v_i, q_i; \theta_{\mathrm{\iota}}^{\mathrm{d}}) = e_{\mathrm{\iota}}^{\mathrm{d}} \left( e_{\mathrm{v}} \left(v_i \right), e_{\mathrm{q}} \left(q_i \right), f_{\mathrm{\iota}} \left(v_i, q_i \right) \right), \label{eq:de-vqa} \\
	& \mathrm{CE}^{\mathrm{d}}_{\mathrm{\iota}} = - \left[ a_i \right] \log\left( \mathrm{softmax} \left( f^{\mathrm{d}}_{\mathrm{\iota}}(v_i, q_i) \right) \right). \label{eq:de-loss-vqa}
\end{align}

Nevertheless, these debiasing methods obtain high out-of-distribution performance at the sacrifice of in-distribution performance. The above issues faced by non-debiasing and debiasing methods raise the question of what causes the current situations and whether we can make the best of both worlds.

\subsubsection{Adaptive Loose Optimization} The parameters of debiasing methods are updated by minimizing the loss function $\mathrm{CE}^{\mathrm{d}}_{\mathrm{\iota}}$ to weaken the bias learning, such as decreasing the connections between critical words of questions and answers. In contrast, for non-debiasing methods, minimizing the loss function $\mathrm{CE}_{\mathrm{\iota}}$ will strengthen the connections. For instance, in Fig. \ref{fig:vqa-arc-comp} (a), a loss of $-\log 0.3 \approx 0.52$ will still be applied to optimize the method, even if \emph{red} is predicted correctly over a wide answer space (typically 3,129 for VQA v2). Therefore, this excessively rigorous optimization would force non-debiasing methods to memorize strong connections between critical words of questions and answers. Similarly, the optimization would excessively prevent debiasing methods from learning the connections, even if slight connections may enhance the QA performance.

The close relationship between the optimization state of models and the loss value is widely acknowledged \cite{garipov2018loss}. In other words, the loss value can serve as an indicator of the state of model parameters. Motivated by this, we propose a novel loss function with adaptive loose optimization $\mathrm{ALO}$, which weakens the bias learning for non-debiasing methods while preserving slight bias learning for debiasing methods. Assuming $loss_{t-1}$ and $loss_{t}$ represent the previous and current loss, respectively, on a mini-batch training data with $loss_{t-1} < loss_{t}$, this indicates a larger gradient will be used to update model parameters in the current optimization compared with the previous one. To avoid the non-debiasing (debiasing) methods to learn (weaken) the bias excessively, \emph{i.e.}, to control the loose degree, we employ the ratio of the last and current loss adaptively. For example, if $loss_{t-1}=0.2$ and $loss_{t}=0.3$, a loss of $- \log 0.3^{\frac{2}{3}} \approx 0.35$ will be applied to update model parameters in the $t$-th optimization. The loss function with adaptive loose optimization for non-debiasing and debiasing methods can be formulated as follows:
\begin{align}
	\mathrm{ALO}_{\mathrm{\iota}} &= - \left[ a_i \right] \log\left( \mathrm{softmax} \left( f_{\mathrm{\iota}} \left(v_i, q_i \right) \right) \right)^{\gamma}, \\
	\mathrm{ALO}^{\mathrm{d}}_{\mathrm{\iota}} &= - \left[ a_i \right] \log\left( \mathrm{softmax} \left( f_{\mathrm{\iota}}^{\mathrm{d}} \left(v_i, q_i \right) \right) \right)^{\gamma}, \\
	\gamma &= \min(\frac{\mathrm{ALO}_{t-1}}{\mathrm{ALO}_{t}}, 0.999), \label{eq:gamma}
\end{align}
where $\mathrm{ALO}_{\mathrm{\iota}}$ denotes the loss after loose optimization for non-debiasing methods, $\mathrm{ALO}^{\mathrm{d}}_{\mathrm{\iota}}$ represents the loss after loose optimization for debiasing methods. We omit $\iota$ and $\mathrm{d}$ for $\mathrm{ALO}_{t}$ for simplicity, where $\mathrm{ALO}_{t}$ denote the loss of the non-debiasing and debiasing QA methods on the $t$-th mini-batch training data. Furthermore, $\gamma$ is the loose degree, and $0.999$ is used to ensure effective loose optimization in the situation where $\frac{\mathrm{ALO}_{t-1}}{\mathrm{ALO}_{t}} > 1$.

\subsection{Extractive QA}
\subsubsection{Task Formulation} Given a dataset $\mathcal{S}_{\mathrm{\kappa}} = \{(p_j, q_j, a_j)\}_{j=1}^{m}$ that consists of $m$ triplets with a passage $p_j$, a natural language question $q_j$, and an answer $a_j$, a system should optimize parameters $\theta_{\kappa}$ of the function $f_{\mathrm{\kappa}}: (p_j, q_j; \theta_{\kappa}) \rightarrow a_j$ to predict answers correctly. For extractive QA, the answer $a_j$ is a single span in the passage $p_j$, and can be denoted as $[a_j^{s}, a_j^{e}]$, where $a_j^{s}$/$a_j^{e}$ denote the start/end span index, $1 \leq a_j^s \leq a_j^e \leq l_{p_j}$, and $l_{p_j}$ represents the length of $p_j$. Following \cite{rajpurkar2016squad, ko2020look, shinoda2022look}, we consider the extractive QA task as a multi-class classification problem.

\subsubsection{Non-debiasing Methods} 
The procedure of current non-debiasing lightweight (small) models for extractive QA is similar to visual QA. The methods \cite{Chen18g, ko2020look} for extractive QA usually use a passage encoder $e_{\mathrm{p}}: p_j \rightarrow \mathbb{R}^{m_{\mathrm{p}} \times d_{\mathrm{p}}}$ like LSTM to learn $m_{\mathrm{p}}$ word-level vectors with $d_{\mathrm{p}}$ dimensions, a question encoder $e_{\mathrm{q}}: q_j \rightarrow \mathbb{R}^{d_{\mathrm{q}}}$ such as GRU to obtain sentence-level vectors with $d_{\mathrm{q}}$ dimensions, a fusion encoder $e_{\mathrm{f}}: \mathbb{R}^{m_{\mathrm{p}} \times d_{\mathrm{p}}} \times \mathbb{R}^{d_{\mathrm{q}}} \rightarrow \mathbb{R}^{m_{\mathrm{p}} \times d_{\mathrm{f}}}$ such as global attention mechanisms to capture the similarity between questions and sentences within passages, and then two separate classifiers $e_{\mathrm{c}}^\mathrm{s}: \mathbb{R}^{m_{\mathrm{p}} \times d_{\mathrm{f}}} \rightarrow  \mathbb{R}^{m_{\mathrm{p}}}$ and $e_{\mathrm{c}}^\mathrm{e}: \mathbb{R}^{m_{\mathrm{p}} \times d_{\mathrm{f}}} \rightarrow \mathbb{R}^{m_{\mathrm{p}}}$ to predict the start and end position respectively. The architecture is shown in Fig. \ref{fig:qa-arc-comp} (a) and can be formulated as follows:
\begin{figure}[tbp]
	\centering  
	\includegraphics[scale=0.5]{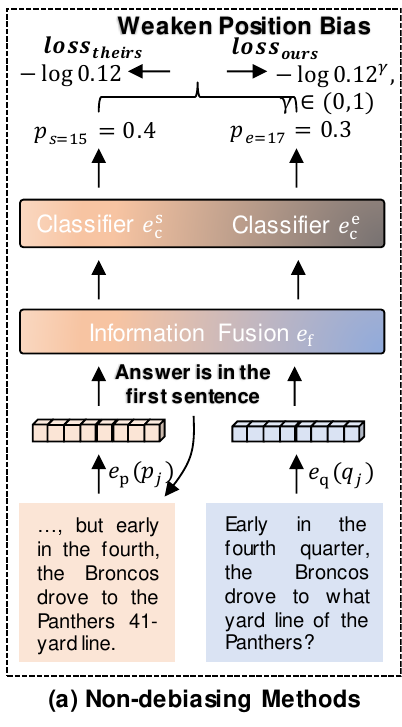}
	\includegraphics[scale=0.5]{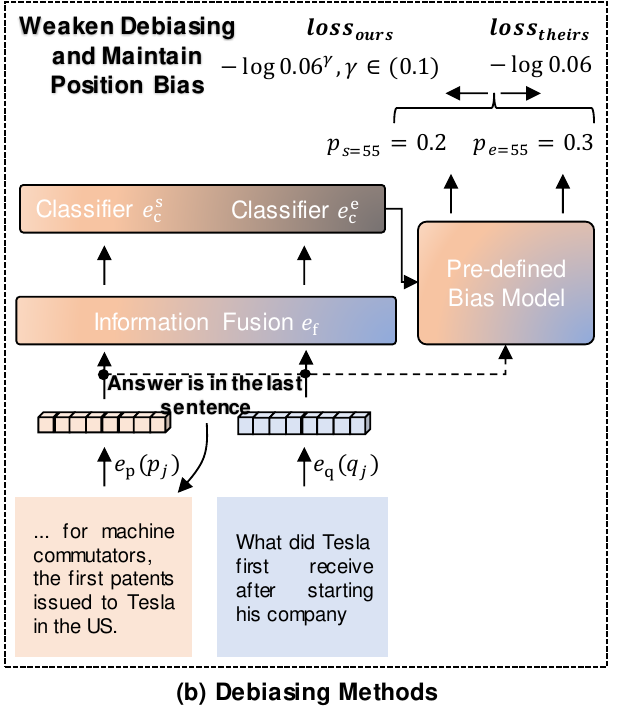}
	\caption{Comparison of non-debiasing and debiasing extractive QA methods with or without adaptive loose optimization.}
	\label{fig:qa-arc-comp}
\end{figure}
\begin{align}
	f_{\mathrm{\kappa}}^{\mathrm{s}} \left(p_j, q_j; \theta_{\kappa} \right) = e_{\mathrm{c}}^{\mathrm{s}} \left(e_{\mathrm{f}} \left( e_{\mathrm{p}} \left(p_j \right), e_{\mathrm{q}} \left(q_j \right) \right) \right), \\
	f_{\mathrm{\kappa}}^{\mathrm{e}} \left(p_j, q_j; \theta_{\kappa} \right) = e_{\mathrm{c}}^{\mathrm{e}} \left(e_{\mathrm{f}} \left( e_{\mathrm{p}} \left(p_j \right), e_{\mathrm{q}} \left(q_j \right) \right) \right).
\end{align}%

To optimize the parameters $\theta_{\kappa}$ of $f_{\mathrm{\kappa}}$, two standard cross-entropy loss functions $\mathrm{CE}_{\mathrm{\kappa}}^\mathrm{s}$ and $\mathrm{CE}_{\mathrm{\kappa}}^\mathrm{e}$ are minimized jointly:
\begin{align}
	\mathrm{CE}_{\mathrm{\kappa}}^\mathrm{s} &= - \left[ a_j^\mathrm{s} \right] \log\left( \mathrm{softmax} \left( f_{\mathrm{\kappa}}^{\mathrm{s}} \left(p_j, q_j \right) \right) \right), \\
	\mathrm{CE}_{\mathrm{\kappa}}^\mathrm{e} &= - \left[ a_j^\mathrm{e} \right] \log\left( \mathrm{softmax} \left( f_{\mathrm{\kappa}}^{\mathrm{e}} \left(p_j, q_j \right) \right) \right),
\end{align}
where $[a_j^\mathrm{s}]$/$[ a_j^\mathrm{e}]$ denote the start/end index of the ground-truth answer $a_j$ respectively.

Non-debiasing methods achieve significantly impressive in-distribution performance. Nevertheless, Ko et al. \cite{ko2020look} found that this success may be obtained by position bias. In other words, they locate answers by the statistical regularity of positional cues, resulting in poor out-of-distribution performance.

\subsubsection{Debiasing Methods} Debiasing methods \cite{ko2020look, he2019unlearn} for extractive QA shown in Fig. \ref{fig:qa-arc-comp} (b) combine the prediction probability from a pre-defined bias model $f_{\mathrm{b}}$ and a vanilla QA model $f_{\kappa}$ to reduce the impact of position bias. To optimize parameters $\theta^{\mathrm{d}}_{\mathrm{\kappa}}$ of the debiasing method $f_{\mathrm{\kappa}}^{\mathrm{d}}$, these methods minimize two standard cross-entropy loss functions $\mathrm{CE}_{\mathrm{\kappa}}^{\mathrm{d}}$ jointly. This process can be formulated as follows:
\begin{align}
	& f^{\mathrm{d}}_{\mathrm{\kappa}}(p_i, q_i; \theta_{\mathrm{\kappa}}^{\mathrm{d}}) = e_{\mathrm{\kappa}}^{\mathrm{d}} \left( f_{\mathrm{b}}(p_i, q_i),	f_{\mathrm{\kappa}} \left(p_i, q_i \right) \right), \\
	& \mathrm{CE}_{\mathrm{\kappa}}^{\mathrm{d}} = - \left[ a_j \right] \log\left( \mathrm{softmax} \left( f_{\mathrm{\kappa}}^{\mathrm{d}} \left(p_j, q_j \right) \right) \right),
\end{align}
where $e_{\kappa}^{\mathrm{d}}$ is the combination strategy such as Learned-Mixin in \cite{ko2020look, ClarkYZ19}, $\mathrm{CE}_{\mathrm{\kappa}}^{\mathrm{d}}$ consists of $\mathrm{CE}_{\mathrm{\kappa}}^{\mathrm{ds}}$ (loss of the start index prediction) and $\mathrm{CE}_{\mathrm{\kappa}}^{\mathrm{de}}$ (loss of the end index prediction), and $[a_j]$ consists of $[a_j^{\mathrm{s}}]$ and $[a_j^{\mathrm{e}}]$.

\subsubsection{Adaptive Loose Optimization} The debiasing methods for extractive QA also achieve high out-of-distribution performance at the expense of in-distribution performance, much like the situations for debiasing visual QA methods. Therefore, we apply the proposed loss function with adaptive loose optimization to weaken the position bias learning for non-debiasing methods $f_{\mathrm{\kappa}}$ while maintaining slight bias learning for debiasing methods $f_{\mathrm{\kappa}}^{\mathrm{d}}$:
\begin{align}
	\mathrm{ALO}_{\mathrm{\kappa}} &= - \left[ a_j \right] \log\left( \mathrm{softmax} \left( f_{\mathrm{\kappa}} \left(p_j, q_j \right) \right) \right)^{\gamma}, \\
	\mathrm{ALO}^{\mathrm{d}}_{\mathrm{\kappa}} &= - \left[ a_j \right] \log\left( \mathrm{softmax} \left( f_{\mathrm{\kappa}}^{\mathrm{d}} \left(p_j, q_j \right) \right) \right)^{\gamma},
\end{align}
where $\gamma$ is the loose degree and can be obtained by Equation \eqref{eq:gamma}.

\section{Experiments}
\label{sec:exp}
We incorporate the proposed loss function into existing non-debiasing and debiasing QA methods and conduct experiments in both in- and out-of-distribution scenarios to validate their effectiveness. Moreover, we carry out a theoretical analysis to bolster the claim of our approach. 
\subsection{Visual QA}
\subsubsection{Datasets} We aim to verify whether our approach can make the best of both worlds in in- and out-of-distribution scenarios. We choose the visual QA datasets from this perspective. 
\begin{itemize}
	\item In-distribution: VQA v2.0 \cite{goyal2017making}. This dataset is a more balanced dataset compared with VQA v1.0. To alleviate language priors, the authors add two complementary but similar images to each question in VQA v1, making the answer of this question different. However, this can only balance the distribution of answers instead of changing the answer distribution between train and test splits. Both VQA v1.0 and v2.0 are split into the following parts: training, validation, test-dev, test-standard, test-challenge, and test-reserve, where the results of test splits are obtained by running on an online server\footnote{\url{https://eval.ai/web/challenges/challenge-page/830/overview}}. Current works \cite{antol2015vqa, yang2022vision, gao2019dynamic, wang2023learning} are typically evaluated and compared based on four perspectives, including their performance in answering \emph{Yes/No} (Y/N) (546K questions in total), \emph{Number} (Num) (415K questions in total), \emph{Other} (144K questions in total), and overall questions.
	\item Out-of-distribution: VQA-CP v1, VQA-CP v2 \cite{agrawal2018don}, and GQA-OOD \cite{kervadec2021roses}. VQA-CP v1 and VQA-CP v2 are developed by re-organizing the training and validation splits of VQA v1.0 and VQA v2.0 respectively. The answer distribution for per question type in the training split is different from the test split. For instance, \emph{1} and \emph{2} are the most common answers for \emph{How many} questions in the VQA-CP v1 training split, while \emph{2} and \emph{4} are the most frequent answers in the test split. Consequently, VQA-CP is typically employed to assess whether methods provide answers by the memorization of statistical regularities. \emph{However, this dataset lacks the validation split to tune hyper-parameters.} Compared with the above datasets, GQA-OOD makes it possible to assess the generalization ability of methods to uncommon or unseen question-answer pairs as well as whether they have absorbed the trends from training data. The GQA-OOD test set consists of \emph{tail} and \emph{head} splits. Specifically, the former is used to evaluate the performance in out-of-distribution scenarios, while the latter is employed to assess the performance in in-distribution situations.
\end{itemize}
\subsubsection{Evaluation Metrics} 
Following \cite{niu2021counterfactual,niu2021introspective}, we employ the harmonic mean of accuracy to evaluate the performance. The calculation is shown in the supplementary material. The details of datasets and metrics are shown in Table \ref{tab:dataset}.
\begin{table}[tbp]
    \centering
    \caption{Details of datasets and evaluation metrics. OEA: open-ended accuracy. SDA: standard accuracy. ID: in-distribution. OOD: out-of-distribution.}
    \label{tab:dataset}
    \resizebox{0.85\columnwidth}{!}{
        \begin{tabular}{ccccccc}
		\toprule
		\textbf{Dataset}  & \textbf{Train} & \textbf{Val} & \textbf{Test} & \textbf{Metric} & \textbf{ID} & \textbf{OOD} \\ \midrule
		VQA v2.0 \cite{goyal2017making} & 443K           & 214K         & 448K          & OEA             & \Checkmark  &  \XSolidBrush   \\ 
		VQA-CP v1 \cite{agrawal2018don}& 245K           & NA           & 125K          & OEA             & \XSolidBrush&  \Checkmark     \\ 
		VQA-CP v2 \cite{agrawal2018don}& 438K           & NA           & 220K          & OEA             & \XSolidBrush&  \Checkmark   \\ 
		GQA-OOD  \cite{kervadec2021roses}& 94K            & 5K           & 2K            & SDA              & \Checkmark  & \Checkmark   \\ 
		\bottomrule
    \end{tabular}
    }
\end{table}

\subsubsection{Baselines} We take the previous state-of-the-art non-debiasing and debiasing methods as baselines to verify whether our approach can make them perform well in in- and out-of-distribution situations simultaneously. 
\begin{itemize}
	\item Non-debiasing: BUTD \cite{anderson2018bottom} and SMRL \cite{cadene2019murel}. In particular, BUTD first combines bottom-up and top-down mechanisms to compute attention weights at the level of objects and then fuses the attended image feature and the question feature to perform answer prediction. SMRL is a simple and fast variant of MUREL \cite{cadene2019murel}, which performs reasoning based on relational modeling between regions. \emph{It is noteworthy that both of them are usually used as backbones in the debiasing methods.}
	\item Debiasing: RUBi \cite{rubi}, CF \cite{niu2021counterfactual}, and CF variant. Specifically, RUBi is a simple model-agnostic learning strategy, which captures language bias by a question-only branch. CF considers language bias to be the direct causal effect of questions on answers, and it eliminates bias by subtracting the direct language effect from the total causal effect. CF variant is a version that removes the visual branch in CF. 
\end{itemize}
\subsubsection{Implementation Details} Our implementations such as data pre-processing and hyper-parameters are the same as CF\footnote{\url{https://github.com/yuleiniu/cfvqa/blob/master/README.md}} for a fair comparison. Specifically, all the methods apply the object/region-based features\footnote{\url{https://github.com/hengyuan-hu/bottom-up-attention-vqa}} as image features and use Skip-Thought vectors\footnote{\url{https://github.com/ryankiros/skip-thoughts}} to initialize the parameters of embedding layers. We set the maximum question length to 14. To obtain the predictions, all of them employ a linear multi-layer perceptron (FC(2048)-FC(1024)-FC(1024)-FC(\emph{answer space})) to project fusion features into the answer space (VQA v2.0 and VQA-CP: 3000, GQA-OOD: 1835). Following \cite{niu2021counterfactual, niu2021introspective}, we restrict the pool of potential answers to only those that appeared no less than nine times in both the training and validation datasets resulting in the mentioned answer space. To train these methods, we apply the Adamax optimizer with an initial learning rate $lr=3e^{-4}$. The learning rate is decayed by 0.25 in the 2-th, and 14-th epochs respectively. We warm up the learning rate within the first seven training epochs. All the methods are trained over 22 epochs with a fixed seed of 1337 and a batch size of 256. The maximum norm of gradients is set to 0.25. For VQA v2.0 and GQA-OOD, we save the model that achieves the best accuracy on the validation split and tested it on the test-dev and test split respectively. The other details are shown in the released code.

\begin{table}[tbp]
        \centering
	\caption{Validation performance (accuracy) comparison of visual QA methods (\%) in in-distribution (VQA v2.0 and GQA-OOD head) and out-of-distribution scenarios (GQA-OOD tail). NDE and DE represent non-debiasing and debiasing methods respectively. HM denotes the harmonic mean of the overall accuracy. In DE methods, the backbone of the upper row is SMRL, while the backbone of the lower row is BUTD.} \label{tab:vqa-val}
	\resizebox{\columnwidth}{!}{
        \begin{tabular}{crcccccccc}
		\toprule
		&  & \multicolumn{4}{c}{\textbf{VQA v2.0 val}} & \multicolumn{3}{c}{\textbf{GQA-OOD val}} & \\ \cmidrule(lr){3-6}	\cmidrule(lr){7-9}
		\multirow{-2}{*}{\textbf{Type}} & \multirow{-2}{*}{\textbf{Methods}} & \textbf{All} & \textbf{Y/N} & \textbf{Num} & \textbf{Other}  & \textbf{All}& \textbf{Tail} & \textbf{Head} & \multirow{-2}{*}{\textbf{HM}} \\ \midrule
  
		& SMRL \cite{cadene2019murel}        & 63.08   & 81.96& 45.68  & 53.38  & 54.57  & 42.97  & 60.44 & 58.52  \\
		& +ALO                               & \textbf{63.36} & 82.02  & 45.95  & 53.79  & \textbf{54.72} & 43.55  & 60.38    & \textbf{58.72}  \\
		& BUTD \cite{anderson2018bottom}     & 64.01  & 82.80 & 44.57  & 59.51  & 55.87  & 44.59          & 61.58  & 59.66 \\
		\multirow{-4}{*}{NDE} & +ALO         & \textbf{64.10} & 82.68  & 44.66   & 55.16  & \textbf{56.52} & 44.56  & 62.58  & \textbf{60.07} \\ \cmidrule(r){1-10}
  
		& RUBi \cite{rubi} & 61.27          & 81.85 & 45.09  & 49.91   & 54.43 & 42.57  & 60.43   & 57.65  \\
		& +ALO             & \textbf{63.29} & 82.12 & 45.74  & 53.65   & \textbf{55.40} & 43.44   & 61.46  & \textbf{59.08} \\
		& CF Variant \cite{niu2021counterfactual}   & 60.76  & 81.54   & 43.38          & 49.61   & 45.82  & 35.83  & 50.89  & 52.24 \\
		& +ALO  &  \textbf{61.44} & 81.93   & 44.42 & 50.37  & \textbf{51.29} & 40.72   & 56.64   & \textbf{55.91} \\
		& CF \cite{niu2021counterfactual}   & 61.11  & 81.20 & 44.76   & 50.14          & 49.96   & 39.59  & 55.21  & 54.98  \\
		& +ALO   & 60.99 &  81.13  & 44.18  & 50.17 & \textbf{51.05} & 40.48   & 56.41   & \textbf{55.58} \\ \cmidrule(r){2-10}
		& RUBi \cite{rubi} & 62.97   & 82.74   & 44.00   & 53.04    & 56.13  & 45.01   & 61.77       & 59.35 \\
		& +ALO    & \textbf{64.24}   & 82.76   & 44.87   & 55.36    & \textbf{56.57}   & 44.58       & 62.65 & \textbf{60.16}      \\
		& CF Variant \cite{niu2021counterfactual} & 63.61    & 82.78   & 44.08         & 54.26       & 56.03 & 45.12 & 61.56 & 59.58 \\
		& +ALO  & \textbf{63.68} & 82.80    & 44.31  & 54.29 & \textbf{56.39} & 44.71  & 62.30       & \textbf{59.81} \\
		& CF \cite{niu2021counterfactual} & 63.19    & 82.48 & 44.16   & 53.65& 52.29  & 41.34       & 57.84  & 57.23  \\
		\multirow{-12}{*}{DE}      & +ALO & \textbf{64.05}   & 82.62   & 44.06& 57.24  & \textbf{52.35} & 40.91 & 58.15 & \textbf{57.61} \\ \bottomrule         
	\end{tabular}
    }
\end{table}

\subsubsection{Main Results} Table \ref{tab:vqa-val} presents the results on the validation split of VQA v2.0 and GQA-OOD datasets, while Tables \ref{tab:vqa-test} displays the results on the test split of VQA v2.0, GQA-OOD, VQA-CP v1, and VQA-CP v2. Our analysis of the results begins with an examination of the ALO approach on the validation split. The results show that combining ALO with baselines results in improved performance, except for CF+ALO on the validation split of VQA v2.0. 
\begin{table*}[tbp]
    \footnotesize
    \caption{Test performance (accuracy) comparison of visual QA methods (\%) in in-distribution (VQA v2.0 and GQA-OOD head) and out-of-distribution scenarios (GQA-OOD tail, VQA-CP v1 and VQA-CP v2). In DE methods, the backbone of the upper row is SMRL, while the backbone of the lower row is BUTD.} \label{tab:vqa-test}
    \resizebox{\textwidth}{!}{
    \begin{tabular}{@{}crcccccccccccccccc@{}}
        \toprule
        \multirow{2}{*}{\textbf{Type}} & \multirow{2}{*}{\textbf{Methods}} & \multicolumn{4}{c}{\textbf{VQA v2.0 test-dev}}                 & \multicolumn{3}{c}{\textbf{GQA-OOD test}}      & \multicolumn{4}{c}{\textbf{VQA-CP v1 test}}                    & \multicolumn{4}{c}{\textbf{VQA-CP v2 test}}                    & \multirow{2}{*}{\textbf{HM}} \\ \cmidrule(lr){3-6} \cmidrule(lr){7-9} \cmidrule(lr){10-13} \cmidrule(lr){14-17}
                                       &                                   & \textbf{All}   & \textbf{Y/N} & \textbf{Num.} & \textbf{Other} & \textbf{All}   & \textbf{Tail} & \textbf{Head} & \textbf{All}   & \textbf{Y/N} & \textbf{Num.} & \textbf{Other} & \textbf{All}   & \textbf{Y/N} & \textbf{Num.} & \textbf{Other} &                              \\ \midrule
        \multirow{4}{*}{NDE}           & SMRL \cite{cadene2019murel}       & 64.76          & 82.20        & 46.44         & 54.01          & 46.32          & 41.67         & 49.16         & 36.86          & 43.39        &                                12.88         & 40.22          & 37.09          & 41.85        & 12.76         & 41.28          & 43.90                        \\
                                       & +ALO                              & \textbf{65.02} & 82.35        & 46.45         & 54.42          & \textbf{47.00} & 40.92         & 50.72         & \textbf{36.86} & 43.53        & 13.56         & 40.20          & \textbf{37.88} & 43.18        & 12.95         & 42.06          & \textbf{44.35}               \\
                                       & BUTD \cite{anderson2018bottom}    & 65.78          & 83.07        & 45.88         & 55.54          & 47.75          & 42.62         & 50.89         & 37.40          & 43.27        & 12.89         & 41.57          & 38.04          & 43.41        & 12.92         & 42.26          & 44.86                        \\
                                       & +ALO                              & \textbf{65.87} & 83.03        & 45.66         & 55.82          & \textbf{48.35} & 43.37         & 51.41         & \textbf{37.62} & 43.91        & 12.98         & 41.76          & \textbf{38.07} & 43.38        & 12.69         & 42.26          & \textbf{45.09}               \\ \midrule
        \multirow{12}{*}{DE}           & RUBi \cite{rubi}                  & 63.28          & 82.28        & 45.46         & 51.05          & 46.78          & 42.52         & 49.39         & 50.83          & 80.18        &                                16.52         & 39.43          & 47.61          & 74.68        & 20.31         & 43.23          & 51.38                        \\
                                       & +ALO                              & \textbf{65.09} & 82.50        & 46.76         & 54.37          & \textbf{47.25} & 41.02         & 51.07         & \textbf{50.86} & 80.81        & 16.56         & 39.38          & 47.13          & 74.79        & 18.75         & 43.49          & \textbf{51.67}               \\
                                       & CF Variant \cite{niu2021counterfactual} & 62.63    & 82.14        & 44.02         & 50.14          & 39.34          & 35.09         & 41.95         & 43.76          & 60.83        & 13.92         & 38.92          & 54.04          & 88.23        & 30.86         & 42.71          & 48.34                        \\
                                       & +ALO                              & \textbf{63.32} & 82.40        & 45.34         & 51.05          & \textbf{45.10} & 40.08         & 48.18         & \textbf{44.98} & 63.74        & 14.47         & 39.29          & \textbf{55.02} & 89.97        & 33.52         & 42.59          & \textbf{51.04}               \\
                                       & CF \cite{niu2021counterfactual}   & 63.01          & 81.96        & 45.48         & 50.98          & 44.28          & 41.20         & 44.28         & 56.88          & 89.75        & 17.56         & 40.21          & 55.42          & 90.56        & 26.61         & 45.65          & 54.00                        \\
                                       & +ALO                              & \textbf{63.02} & 81.75        & 45.21         & 51.01          & \textbf{44.71} & 40.55         & 47.26         & \textbf{57.37} & 89.91        & 17.33         & 41.10          & 55.29          & 90.61        & 25.69         & 45.69          & \textbf{54.24}               \\ \cmidrule(l){2-18} 
                                       & RUBi \cite{rubi}                  & 64.94          & 83.22        & 45.51         & 53.71          & 48.03          & 42.24         & 51.59         & 50.45          & 80.25        & 14.76         & 41.01          & 39.57          & 49.74        & 19.17         & 42.38          & 49.19                        \\
                                       & +ALO                              & \textbf{66.02} & 83.33        & 45.29         & 55.95          & \textbf{49.18} & 46.00         & 51.13         & \textbf{50.69} & 80.92        & 14.90         & 41.64          & 39.20          & 45.51        & 17.21         & 42.06          & \textbf{49.56}               \\
                                       & CF Variant \cite{niu2021counterfactual} & 65.19    & 82.98        & 44.93         & 54.58          & 48.03          & 44.21         & 50.38         & 37.26          & 44.99        & 13.08         & 41.68          & 37.59          & 44.04        & 13.03         & 41.97          & 44.64                        \\
                                       & +ALO                              & \textbf{65.91} & 83.09        & 45.01         & 55.98          & \textbf{48.86} & 42.71         & 52.63         & 36.78          & 44.48        & 12.90         & 41.56          & \textbf{37.93} & 42.77        & 13.13         & 42.35          & \textbf{44.85}               \\
                                       & CF \cite{niu2021counterfactual}   & 65.47          & 83.16        & 44.72         & 55.07          & 45.24          & 41.11         & 47.78         & 57.64          & 89.18        & 14.57         & 43.75          & 54.02          & 91.35        & 13.46         & 45.60          & 54.62                        \\
                                       & +ALO                              & 65.43          & 82.95        & 45.28         & 55.03          & \textbf{46.21} & 40.55         & 49.68         & \textbf{57.66} & 89.28        & 15.39         & 43.47          & 53.98          & 91.20        & 13.44         & 45.64          & \textbf{54.95}               \\ \bottomrule
    \end{tabular}}
\end{table*}
Next, we evaluate the generalization ability of ALO on the test split. The results indicate that our approach can enhance the performance of baseline models on the test split of both VQA v2.0 and GQA-OOD. In the following subsection, we will further discuss the results obtained on the VQA-CP dataset.

Additionally, from a non-debiasing perspective, it is clear that our proposed approach ALO improves the test performance of non-debiasing methods in both in- and out-of-distribution scenarios. In particular, ALO enhances BUTD by 0.44\% on the test split of both VQA v2.0 and GQA-OOD, according to the HM metric. Furthermore, ALO significantly improves the performance of debiasing methods in these scenarios, particularly CF Variant, which saw a 4.35\% increase on the test split of VQA v2.0 and GQA-OOD, as measured by the HM metric.

Finally, it is worth noting that debiasing strategies such as RUBi and CF with SMRL fail to improve performance on the test split of VQA v2.0, whereas sustainable improvements are achieved using ALO. Since no validation splits are available, we tuned our hyper-parameters on the test split of VQA-CP following \cite{he2019unlearn, liang2020learning, SeoKFH17}. We can see that our approach obtains comparable results on the test split, particularly improving the performance of the SMRL-based method. In brief, these findings demonstrate that ALO enables both non-debiasing and debiasing methods to achieve robust performance under a range of distributional conditions.

\subsubsection{Discussions}
\label{subsec:vqadis}
We notice that previous state-of-the-art debiasing methods on VQA-CP do not achieve success on the GQA-OOD test split. For instance, CF and its variant do not succeed to improve the head and tail performance. This is surprising because they were designed to overcome the reliance on language bias. Furthermore, the effectiveness of the debiasing strategy is dependent on the backbone being used. For example, it is found that RUBi+BUTD resulted in improvements in the test split of GQA-OOD, while RUBi+SMRL does not. In contrast, our proposed optimization enables existing methods to achieve sustainable improvements on various in- and out-of-distribution scenarios. Notably, our proposed approach of combining ALO with RUBi achieved better performance on the test split of VQA v2.0 and GQA-OOD compared to using BUTD and SMRL alone. This improvement was observed despite the fact that RUBi reduced the capacity of SMRL. Furthermore, \emph{due to the lack of validation splits for VQA-CP, previous methods may only tune hyper-parameters on the test split.} Therefore, we think further debiasing studies may conduct the out-of-distribution experiments on multiple datasets rather than just VQA-CP. 

\begin{figure}[tbp]
	\centering  
	\subfloat[In-distribution situation.]{\includegraphics[scale=0.45]{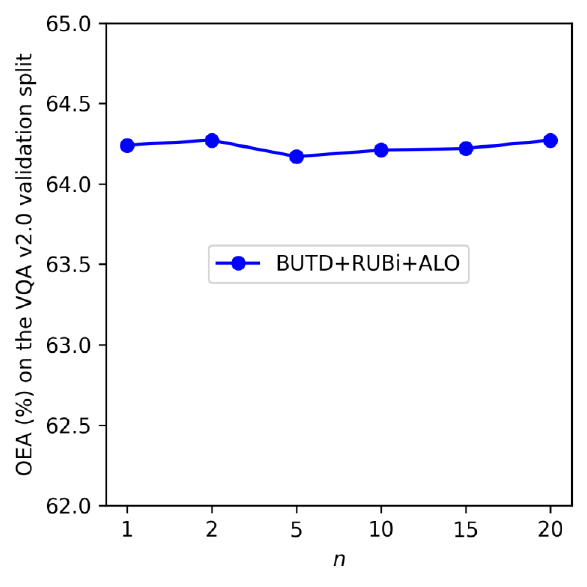}}
	\subfloat[Out-of-distribution Situation.]{\includegraphics[scale=0.45]{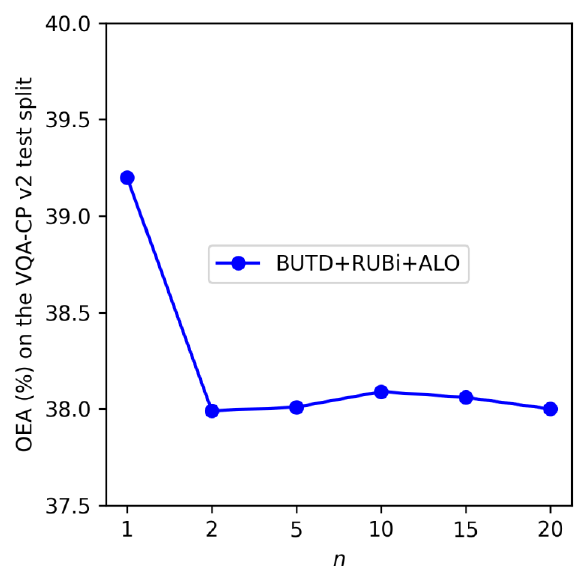}}
	\caption{Impact of batch quantity on adaptive loose optimization. }
	\label{fig:loss-b}
\end{figure}

To explore the impact of batch quantity on adaptive loose optimization, we conduct experiments on the visual QA task. In Equation \eqref{eq:gamma}, we employ the ratio of the previous loss $\mathrm{ALO}_{t-1}$ and the current loss $\mathrm{ALO}_{t}$ to perform loose optimization. Here, we leverage the ratio of $\mathrm{ALO}_{t-n}$ and $\mathrm{ALO}_{t}$ to be the loose degree, where $n$ varies from 1 to 20. We take BUTD+RUBi as an example and conduct experiments on both in- and out-of-distribution scenarios. The results are shown in Fig. \ref{fig:loss-b}. It can be seen that in the in-distribution situation (VQA v2.0 validation split), BUTD+RUBi+ALO obtains slight improvements with the increase of $n$. However, no such trends are observed in the out-of-distribution scenario. It should be noted that ALO with $n=1$ always facilitates the baseline to obtain good performance.

\begin{figure*}[tbp]
    \centering
    \includegraphics[width=0.9\textwidth]{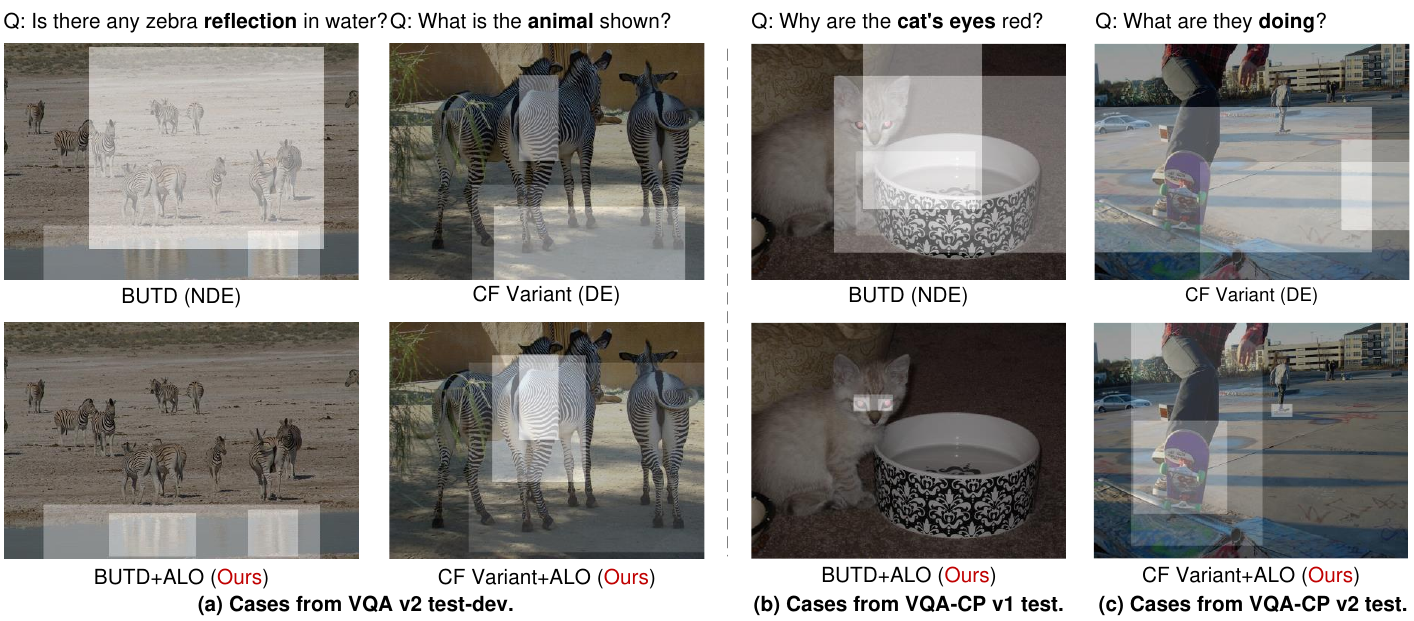}
    \caption{Qualitative results of BUTD and CF Variant and their combination with ALO in in- and out-of-distribution scenarios. We visualize the top three attended regions for each image, wherein the lower opacity denotes the greater attention weight. In-distribution situation: VQA v2. Out-of-distribution situation: VQA-CP v1 and v2.}
    \label{fig:vqa-case_study}
\end{figure*}
\subsubsection{Case Study} In order to qualitatively analyze the effectiveness of ALO in enhancing visual QA methods such as BUTD (Non-debiasing) and CF Variant+BUTD (Debiasing), we visualize the top three attended regions according to their respective attention weight. The weight is obtained by summing the weights of one region attended by all words. These results are presented in Fig. \ref{fig:vqa-case_study}, with regions of greater attention being associated with lower opacity values. For instance, considering the question \emph{Is there any reflection of zebra in water}, which belongs to the in-distribution scenario, our proposed ALO approach enables BUTD to focus explicitly on the visual region \emph{zebra reflection}. Similarly, for a question such as \emph{What are they doing} that belongs to an out-of-distribution situation, our approach enables CF Variant to focus directly on the visual regions \emph{human} and \emph{skate}, thereby reducing noise and increasing the accuracy of predictions. The case studies further reveal that ALO can facilitate both non-debiasing and debiasing methods to focus on critical regions or objects, thereby improving performance in both in- and out-of-distribution scenarios.

\subsection{Extractive QA}
\subsubsection{Datasets}
Without loss of generality, following \cite{niu2021introspective}, we also select the SQuAD variant \cite{ko2020look}, a machine reading comprehension dataset for extractive QA, to verify the effectiveness of adaptive loose optimization in in- and out-of-distribution scenarios. This variant divides the original training split SQuAD\SPSB{total}{train} into several different subsets, which is also applicable to SQuAD\SPSB{total}{dev}. 
\begin{itemize}
	\item In-distribution: training and testing on the same-distribution split. For example, training on the subset SQuAD\SPSB{k=1}{train} where all answers are in the first sentence and testing on SQuAD\SPSB{k=1}{dev}.
	\item Out-of-distribution: biased training and testing. For instance, training on the subset SQuAD\SPSB{k=3}{train} where all answers are in the third sentence and testing on SQuAD\SPSB{k=1}{dev}. \emph{Therefore, experimental results on this dataset can evaluate whether methods suffer from position bias.} The detailed statistics and settings of the SQuAD variant are shown in the supplementary material. 
\end{itemize}

\subsubsection{Evaluation Metrics} To evaluate the QA performance comprehensively, we select the macro-averaged F1 score to be metrics following \cite{rajpurkar2016squad, ko2020look}. The detailed statistics and settings are shown in the supplementary material.

\subsubsection{Baselines} In comparison to BiDAF \cite{SeoKFH17} and XLNet \cite{yang2019xlnet}, Ko et al. \cite{ko2020look} showed that the debiasing strategies including Learned-Mixin (LM) and Bias Product (BP) \cite{hinton2002training}  combined with BERT yield the best F1 score. Therefore, we select BERT as baselines to verify whether loose optimization can make it perform well in in- and out-of-distribution scenarios simultaneously. It is well known that BERT is a large language model. As it is commonly acknowledged, BERT represents a large-scale language model. If our proposed approach can effectively improve the performance of BERT, this would indicate that ALO has the potential for application to other large models, such as ChatGPT.

\subsubsection{Implementation Details} We keep the same settings as BERT+BP and BERT+LM\footnote{\url{https://github.com/dmis-lab/position-bias/blob/master/README.md}} for a fair comparison. Methods are implemented by modifying the open-sourced PyTorch implementations\footnote{\url{https://github.com/huggingface/transformers}}. In particular, the maximum sequence and question length is set to 384 and 64 respectively. The maximum answer length is set to 30. We train BERT+BP+ALO, BERT+LM+ALO, and BERT+ALO for two epochs with a batch size of 12, an initial learning rate such as $3e^{-5}$, and a fixed seed of 42. The maximum norm of gradients is set to 1. The other settings are shown in the released code.

\subsubsection{Main Results} Table \ref{tab:eqa} shows the F1 score comparison in various in- and out-of-distribution situations. First, in the in-distribution situation, we can see that our method can obtain sustainable improvements under the setting such as (SQuAD\SPSB{total}{train}, SQuAD\SPSB{total}{dev}~) and (SQuAD\SPSB{total}{train}, SQuAD\SPSB{k$\neq$1}{dev}~). However, our approach fails to enhance the QA performance on the question whose answer is in the first sentence. Second, in the out-of-distribution situation, our method succeeds to improve the performance of BERT+LM (Debiasing) but fails to obtain improvements on BERT (Non-debiasing). Finally, in the perspective of the harmonic mean, our approach can enhance them in various settings. Specifically, it can be seen that our approach improves BERT by 0.94\%. Regarding the debiasing methods, our approach enables them to achieve enhancements in most settings by slightly reducing their debiasing capability, \emph{i.e.}, maintaining slight regularity memorization of position clues. Our approach enhances BERT+LM and BERT+BP by 0.36\% and 0.47\%, respectively. In brief, the results in various settings demonstrate the effectiveness of our approach.
\begin{table*}[tb]
\caption{Performance (F1 score) comparison of BERT-based extractive QA methods (\%) in in- and out-of-distribution scenarios. (SQuAD\SPSB{k=2}{train}, SQuAD\SPSB{k=1}{dev}) denotes methods are trained on the former and tested on the latter. The cells marked with shading represent experiments conducted under OOD settings.} \label{tab:eqa}
\resizebox{\textwidth}{!}{
\begin{tabular}{@{}cccc>{\columncolor{lightgray!50}}c>{\columncolor{lightgray!50}}cc>{\columncolor{lightgray!50}}c>{\columncolor{lightgray!50}}c>{\columncolor{lightgray!50}}cc>{\columncolor{lightgray!50}}c>{\columncolor{lightgray!50}}c>{\columncolor{lightgray!50}}cc@{}}
\toprule
\multirow{2}{*}{\textbf{Type}}  & \multirow{2}{*}{\textbf{Methods}} & \textbf{SQuAD\SPSB{total}{train}} & \textbf{SQuAD\SPSB{k=1}{train}}   & \textbf{SQuAD\SPSB{k=2}{train}}   & \textbf{SQuAD\SPSB{k=3}{train}}   & \textbf{SQuAD\SPSB{total}{train}} & \textbf{SQuAD\SPSB{k=1}{train}}   & \textbf{SQuAD\SPSB{k=2}{train}}   & \textbf{SQuAD\SPSB{k=3}{train}}   & \textbf{SQuAD\SPSB{total}{train}} & \textbf{SQuAD\SPSB{k=1}{train}}   & \textbf{SQuAD\SPSB{k=2}{train}}   & \textbf{SQuAD\SPSB{k=3}{train}}   & \multirow{2}{*}{\textbf{HM}} \\ \cmidrule(l){3-6} \cmidrule(l){7-10} \cmidrule(l){11-14}
                      & & \multicolumn{4}{c}{\textbf{SQuAD\SPSB{k=1}{dev}}}  & \multicolumn{4}{c}{\textbf{SQuAD\SPSB{k$\neq$1}{dev}}}         & \multicolumn{4}{c}{\textbf{SQuAD\SPSB{total}{dev}}}         \\ \midrule
\multirow{2}{*}{NDE}       & NA                       & 89.31        & 86.67 & 31.99 & 21.38 & 88.55        & 15.97 & 47.78 & 71.90  & 88.81        & 40.30  & 42.35 & 54.52 & 41.73            \\
                      & +ALO                     & 89.25        & 86.59 & \textbf{33.40}  & \textbf{24.94} & \textbf{88.75}        & 15.77 & 46.24 & 71.37 & \textbf{88.92}        & 40.13 & 41.83 & \textbf{55.39} & \textbf{42.66}            \\ \midrule
\multirow{4}{*}{DE}       & BP \cite{hinton2002training}     & 88.92        & 86.85 & 37.44 & 33.48 & 88.66        & 21.07 & 56.95 & 78.80  & 88.75        & 43.7  & 50.24 & 63.21 & 50.37            \\
                      & +ALO               & 88.75   & \textbf{87.56} & \textbf{41.43} & \textbf{35.27} & \textbf{89.03}  & 19.80  & \textbf{58.25} & 78.61 & \textbf{89.12}        & 43.04 & \textbf{51.21} & \textbf{63.71} & \textbf{50.74}            \\
                    & LM \cite{ClarkYZ19}      & 88.77        & 85.33 & 81.00    & 77.45 & 88.50         & 78.82 & 79.71 & 78.43 & 88.60         & 81.06 & 80.16 & 78.10  & 81.95            \\
                      & +ALO                     & 88.59        & \textbf{85.90}  & \textbf{81.67} & \textbf{78.38} & \textbf{88.70}         & \textbf{79.18} & \textbf{80.09} & \textbf{79.08} & \textbf{88.66}        & \textbf{81.49} & \textbf{80.64} & \textbf{78.84} & \textbf{82.41}            \\ \bottomrule
\end{tabular}}
\end{table*}

\subsubsection{Discussions} The combination of LM (BP) and BERT exhibits slightly poorer performance than that of BERT alone under in-distribution scenarios, as demonstrated by the findings outlined in Table \ref{tab:eqa}. In other words, they originally performed poorly in answering the first sentence question. This may be the reason for the decline in our method. However, we can see that, with the combination of ALO and BERT, this dilemma is alleviated effectively. Moreover, we note that the variant of SQuAD lacks the test split to evaluate the generalization ability. This observation motivates us to develop a benchmark dataset that is capable of evaluating robustness comprehensively in the future. 

\begin{figure}[tbp]
	\centering  
	\subfloat[(SQuAD\SPSB{k=1}{train}, SQuAD\SPSB{k=1}{dev}).]{\includegraphics[scale=0.45]{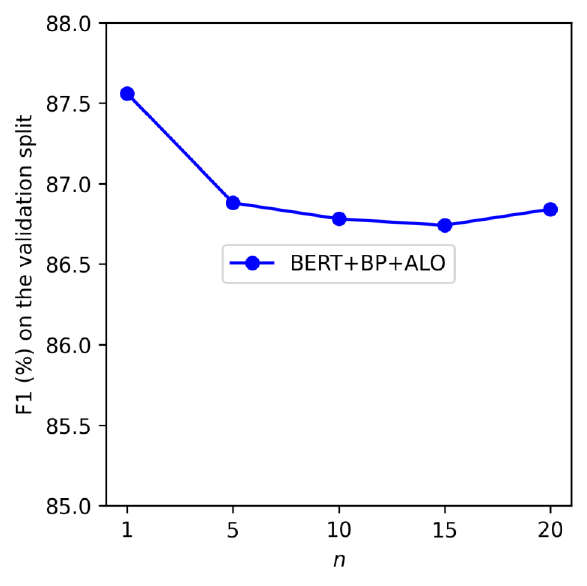}}
	\subfloat[(SQuAD\SPSB{k=1}{train}, SQuAD\SPSB{k=2}{dev}).]{\includegraphics[scale=0.45]{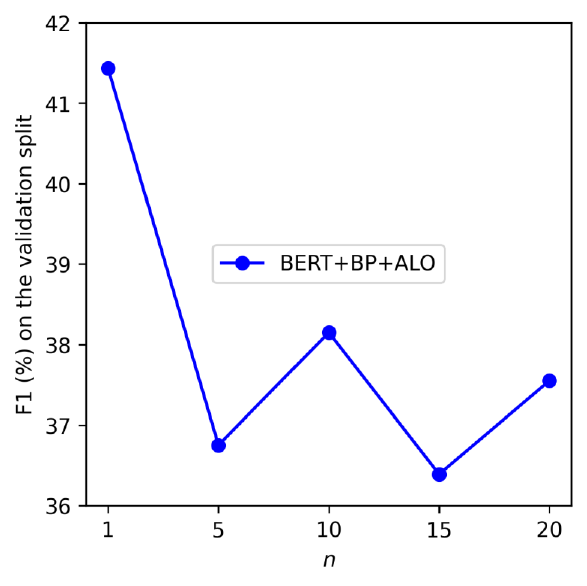}}
	\caption{Impact of batch quantity on adaptive loose optimization. The left setting is in-distribution while the right is out-of-distribution.}
	\label{fig:loss-b-qa}
\end{figure}

Similar to the experiments in Section \ref{subsec:vqadis}, we investigate the impact of batch quantity on adaptive loose optimization on the extractive QA task. We take BERT+BP as an example and conduct experiments on both in-distribution (SQuAD\SPSB{k=1}{train}, SQuAD\SPSB{k=1}{dev}) and out-of-distribution situations (SQuAD\SPSB{k=1}{train}, SQuAD\SPSB{k=2}{dev}). The results are shown in Fig. \ref{fig:loss-b-qa}. It is observed that the results deteriorate as $n$ increases, possibly due to ineffective loose optimization. Specifically, $\mathrm{ALO}_{t-n}$ may be smaller than $\mathrm{ALO}_{t}$ with a large $n$, which will lead to $\gamma$ in Equation \eqref{eq:gamma} being 0.999. Based on the findings depicted in Fig. \ref{fig:loss-b} and \ref{fig:loss-b-qa}, we conclude that setting a small value for $n$ is crucial for effective adaptive loose optimization.

\subsubsection{Case Study} 
\begin{figure*}[tbp]
	\centering  
	\subfloat[In-distribution: (SQuAD\SPSB{total}{train}, SQuAD\SPSB{k=1}{dev}).]{\includegraphics[width=\textwidth]{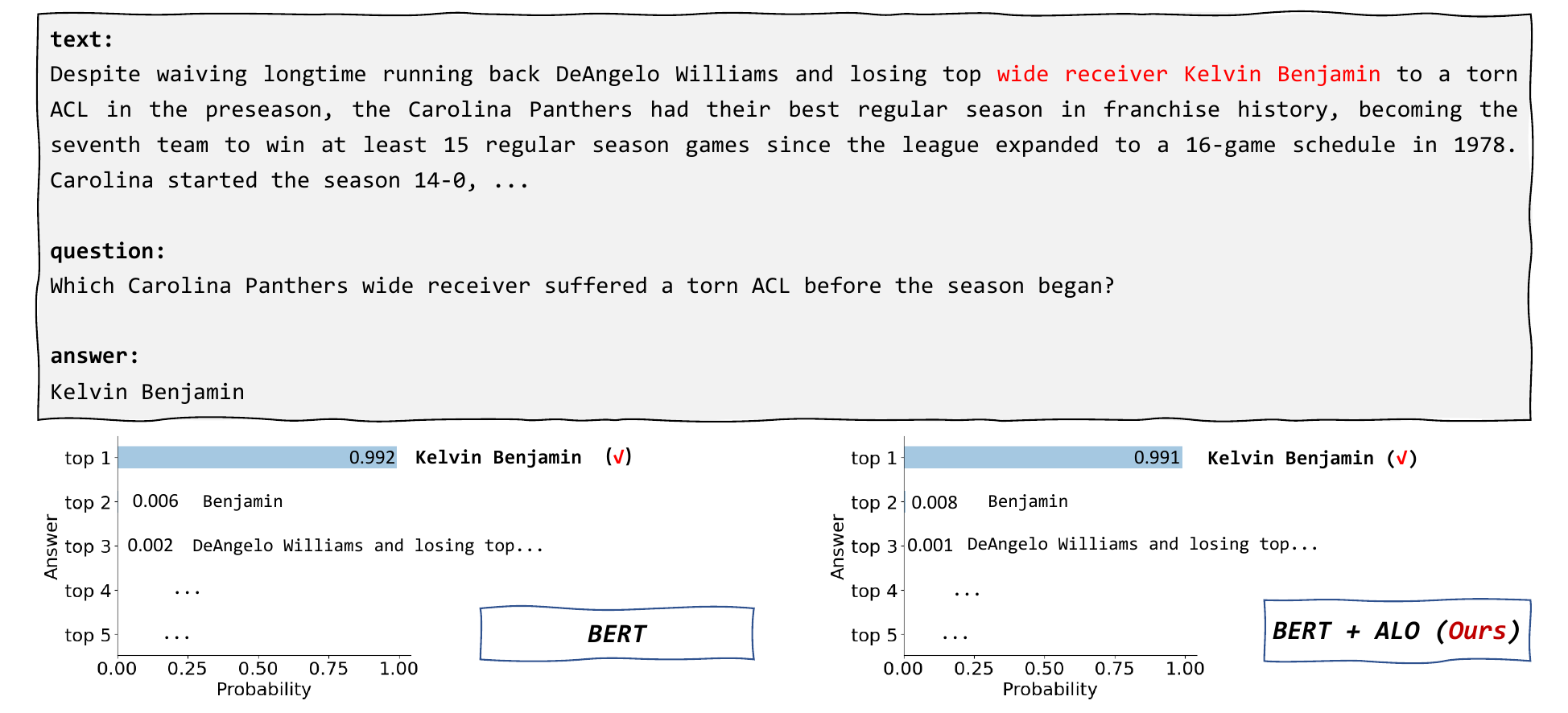}}
	\hspace{1cm}
	\subfloat[Out-of-distribution: (SQuAD\SPSB{k=2}{train}, SQuAD\SPSB{k=1}{dev}).]{\includegraphics[width=\textwidth]{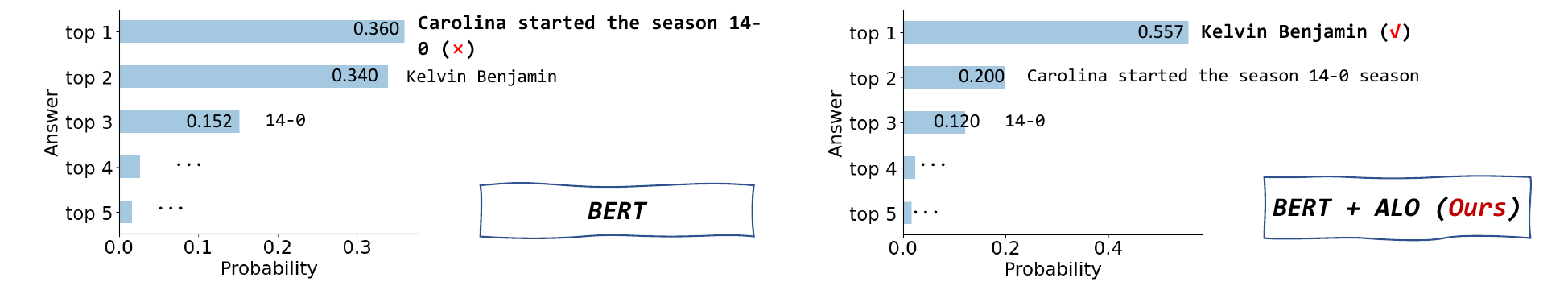}}
	\caption{Qualitative results of BERT (non-debiasing method) and its combination with ALO in in- and out-of-distribution situations.}
	\label{fig:qa-case_study2}
\end{figure*}

\begin{figure*}[tbp]
	\centering  
	\subfloat[In-distribution: (SQuAD\SPSB{total}{train}, SQuAD\SPSB{k=3}{dev}).]{\includegraphics[width=\textwidth]{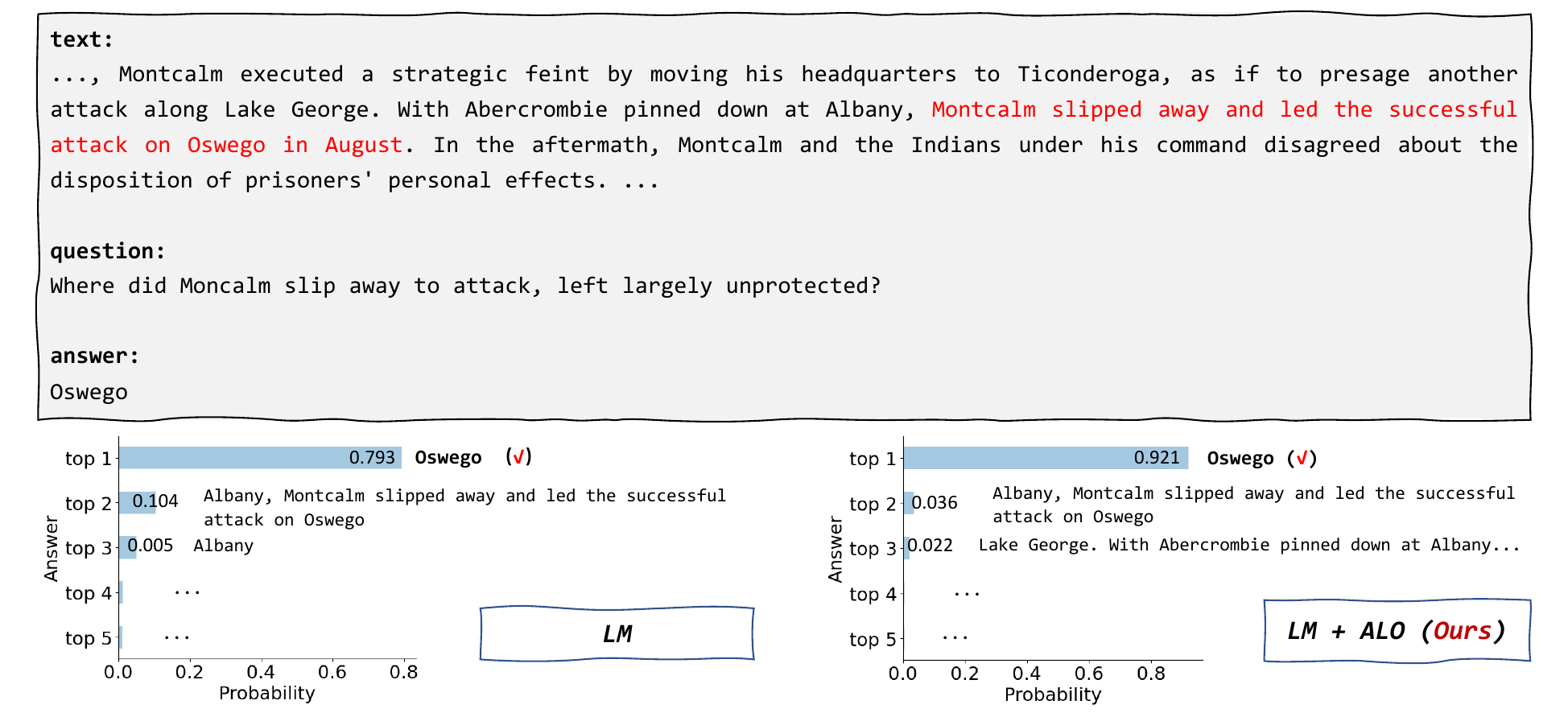}}
	\hspace{1cm}
	\subfloat[Out-of-distribution: (SQuAD\SPSB{k=1}{train}, SQuAD\SPSB{k=3}{dev}).]{\includegraphics[width=\textwidth]{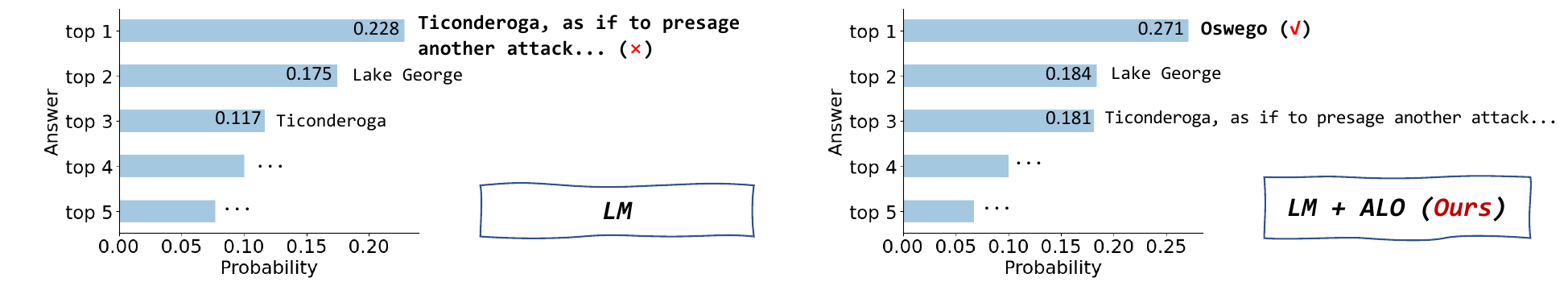}}
	\caption{Qualitative results of BERT+LM (debiasing method) and its combination with ALO in in- and out-of-distribution situations.}
	\label{fig:qa-case_study1}
\end{figure*}

The results in Fig. \ref{fig:qa-case_study2} and \ref{fig:qa-case_study1} are provided to qualitatively analyze whether our approach can facilitate BERT and BERT+LM, \emph{i.e.}, non-debiasing and debiasing methods, to perform well in both in- and out-of-distribution scenarios. On one head, for non-debiasing methods, we designate the settings of (SQuAD\SPSB{total}{train}, SQuAD\SPSB{k=1}{dev}) and (SQuAD\SPSB{k=2}{train}, SQuAD\SPSB{k=1}{dev}) as the in- and out-of-distribution situations, respectively, where ``(\#arg1, \#arg2)" denotes methods are trained on the former and tested on the latter. On the other hand, for debiasing methods, we choose the settings of (SQuAD\SPSB{total}{train}, SQuAD\SPSB{k=3}{dev}) and (SQuAD\SPSB{k=1}{train}, SQuAD\SPSB{k=3}{dev}) to be the in- and out-of-distribution scenarios, respectively. We visualize the probability of the top three predicted answers. It can be seen that our approach can facilitate the above methods to perform well in the mentioned scenarios. Specifically, in the in-distribution situation, our method achieves a comparable or higher predicted answer probability compared with baselines, although both of them accurately answer the question.
In the out-of-distribution scenario, ALO can facilitate baselines to predict answers accurately, while the baseline alone cannot.

\subsection{Theoretical Analysis}
We will discuss why the adaptive loose optimization for QA methods can achieve robust performance from the perspective of loss functions' derivatives. Taking loose optimization for debiasing methods as examples, their loss functions usually have two parts: one is the branch loss such as a question-only branch in RUBi \cite{rubi}, and the other is the debiased loss in Equation (\ref{eq:de-loss-vqa}). As is known to all, the derivatives of a multiclass cross-entropy function $\mathrm{CE}$ with respect to outputs $z$ are as follows:
\begin{equation}
	\begin{split}
		\frac{\partial \mathrm{CE}}{\partial z} = \frac{\partial \mathrm{CE}}{\partial \hat{a}} \frac{\partial \hat{a}}{\partial z} = a(1-p),
	\end{split}
\end{equation}
where $\hat{a} = \mathrm{softmax}(z)$, $a \in \{\pm1\}$ denotes whether an answer is true for a specific question, $p$ is the prediction probability for answers. Therefore, we obtain the derivative division of loss functions with respect to outputs:
\begin{align}
	\frac{ \frac{\partial \mathcal{L}_{\mathrm{ours}} }{\partial z} }{ \frac{\partial \mathcal{L}_{\mathrm{theirs}} }{\partial z} } = \frac{ a(1-p_{\mathrm{b}}) + a(1-p_{\mathrm{d}}^{\gamma})}{a(1-p_{\mathrm{b}}) + a(1-p_{\mathrm{d}})} \leq 1, \label{eq:theo-anal}
\end{align}
where $p_{\mathrm{b}}$ is the probability predicted by a single branch such as a question-only branch in RUBi \cite{rubi}, and $p_{\mathrm{d}}$ is the probability obtained by $\mathrm{softmax}(f^{\mathrm{d}}_{\mathrm{\iota}}(v_i, q_i))$ (Please see details in Equation (\ref{eq:de-vqa})). 

We can obtain $(1-p_{\mathrm{d}}^{\gamma}) \leq ( 1-p_{\mathrm{d}})$ owing to $0 < \gamma < 1$ and $0 \leq p_{\mathrm{d}} \leq 1$. Therefore, we can further get the upper bound of Equation (\ref{eq:theo-anal}) is 1. This shows that our approach will leverage slightly smaller gradients to update training parameters. In other words, the adaptive loose optimization will weaken the debiasing ability and maintain slight bias learning for debiasing methods. Similarly, it will weaken bias learning for non-debiasing methods. In short, we demonstrate the validity of our approach by theoretical analyses and extensive experiments.  

\section{Conclusion}
\label{sec:con}
This paper proposes a novel loss function with adaptive loose optimization, to facilitate existing QA methods to achieve robust performance in in- and out-of-distribution scenarios simultaneously. The primary objective of the adaptive loose optimization is to reduce bias learning, such as connections between critical words of questions and answers in visual QA, for non-debiasing methods while preserving slight bias learning for debiasing methods. Experimental results on VQA v2.0, VQA-CP v1, VQA-CP v2, GQA-OOD datasets, and the variant of SQuAD show that the adaptive loose optimization enables QA methods to achieve the best of both in- and out-of-distribution situations. Based on the findings, we can conclude that it is possible to enhance performance in both distribution settings. Furthermore, the results also demonstrate that our proposed method is easily extensible to other methods.

\section*{Acknowledgments}
This work was supported by the National Key Research and Development Program of China (2021YFB1715600), the National Natural Science Foundation of China (U22B2019, 62272372, 62293553, 62250066, 621737002).

\bibliographystyle{IEEEtran}
\bibliography{references}

\begin{thebibliography}{10}
\providecommand{\url}[1]{#1}
\csname url@samestyle\endcsname
\providecommand{\newblock}{\relax}
\providecommand{\bibinfo}[2]{#2}
\providecommand{\BIBentrySTDinterwordspacing}{\spaceskip=0pt\relax}
\providecommand{\BIBentryALTinterwordstretchfactor}{4}
\providecommand{\BIBentryALTinterwordspacing}{\spaceskip=\fontdimen2\font plus
\BIBentryALTinterwordstretchfactor\fontdimen3\font minus \fontdimen4\font\relax}
\providecommand{\BIBforeignlanguage}[2]{{%
\expandafter\ifx\csname l@#1\endcsname\relax
\typeout{** WARNING: IEEEtran.bst: No hyphenation pattern has been}%
\typeout{** loaded for the language `#1'. Using the pattern for}%
\typeout{** the default language instead.}%
\else
\language=\csname l@#1\endcsname
\fi
#2}}
\providecommand{\BIBdecl}{\relax}
\BIBdecl

\bibitem{rajpurkar2016squad}
P.~Rajpurkar, J.~Zhang, K.~Lopyrev, and P.~Liang, ``{SQuAD: 100, 000+ Questions for Machine Comprehension of Text},'' in \emph{EMNLP}, 2016, pp. 2383--2392.

\bibitem{hu2019read}
M.~Hu, F.~Wei, Y.~Peng, Z.~Huang, N.~Yang, and D.~Li, ``{Read+ Verify: Machine Reading Comprehension with Unanswerable Questions},'' in \emph{AAAI}, 2019, pp. 6529--6537.

\bibitem{antol2015vqa}
S.~Antol, A.~Agrawal, J.~Lu, M.~Mitchell, D.~Batra, C.~L. Zitnick, and D.~Parikh, ``{VQA: Visual Question Answering},'' in \emph{ICCV}, 2015, pp. 2425--2433.

\bibitem{yanvqa}
H.~Li, J.~Huang, P.~Jin, G.~Song, Q.~Wu, and J.~Chen, ``{Weakly-Supervised 3D Spatial Reasoning for Text-based Visual Question Answering},'' \emph{IEEE TIP}, 2023.

\bibitem{9525040}
J.~Ma, J.~Liu, Q.~Lin, B.~Wu, Y.~Wang, and Y.~You, ``{Multitask Learning for Visual Question Answering},'' \emph{TNNLS}, vol.~34, no.~3, pp. 1380--1394, 2023.

\bibitem{zhang2023toward}
Y.~Han, J.~Yin, J.~Wu, Y.~Wei, and L.~Nie, ``{Semantic-aware Modular Capsule Routing for Visual Question Answering},'' \emph{IEEE TIP}, 2023.

\bibitem{rubi}
R.~Cadene, C.~Dancette, H.~Ben-younes, M.~Cord, and D.~Parikh, ``{RUBi: Reducing Unimodal Biases for Visual Question Answering},'' in \emph{NeurIPS}, 2019, p. 841–852.

\bibitem{niu2021counterfactual}
Y.~Niu, K.~Tang, H.~Zhang, Z.~Lu, X.-S. Hua, and J.-R. Wen, ``{Counterfactual VQA: a Cause-effect Look at Language Bias},'' in \emph{CVPR}, 2021, pp. 12\,700--12\,710.

\bibitem{ko2020look}
M.~Ko, J.~Lee, H.~Kim, G.~Kim, and J.~Kang, ``{Look at the First Sentence: Position Bias in Question Answering},'' in \emph{EMNLP}, 2020, pp. 1109--1121.

\bibitem{hinton2002training}
G.~E. Hinton, ``{Training Products of Experts by Minimizing Contrastive Divergence},'' \emph{Neural Computation}, vol.~14, no.~8, pp. 1771--1800, 2002.

\bibitem{cadene2019murel}
R.~Cadene, H.~Ben-Younes, M.~Cord, and N.~Thome, ``{MUREL: Multimodal Relational Reasoning for Visual Question Answering},'' in \emph{CVPR}, 2019, pp. 1989--1998.

\bibitem{goyal2017making}
Y.~Goyal, T.~Khot, D.~Summers-Stay, D.~Batra, and D.~Parikh, ``{Making the V in VQA Matter: Elevating the Role of Image Understanding in Visual Question Answering},'' in \emph{CVPR}, 2017, pp. 6904--6913.

\bibitem{selvaraju2019}
R.~R. Selvaraju, S.~Lee, Y.~Shen, H.~Jin, S.~Ghosh, L.~Heck, D.~Batra, and D.~Parikh, ``{Taking a Hint: Leveraging Explanations to Make Vision and Language Models More Grounded},'' in \emph{ICCV}, 2019, pp. 2591--2600.

\bibitem{shah2019cycle}
M.~Shah, X.~Chen, M.~Rohrbach, and D.~Parikh, ``{Cycle-consistency for Robust Visual Question Answering},'' in \emph{CVPR}, 2019, pp. 6649--6658.

\bibitem{ClarkYZ19}
C.~Clark, M.~Yatskar, and L.~Zettlemoyer, ``{Don't Take the Easy Way Out: Ensemble Based Methods for Avoiding KnownDataset Biases},'' in \emph{EMNLP}, 2019, pp. 4067--4080.

\bibitem{agrawal2018don}
A.~Agrawal, D.~Batra, D.~Parikh, and A.~Kembhavi, ``{Don't Just Assume; Look and Answer: Overcoming Priors for Visual Question Answering},'' in \emph{CVPR}, 2018, pp. 4971--4980.

\bibitem{kervadec2021roses}
C.~Kervadec, G.~Antipov, M.~Baccouche, and C.~Wolf, ``{Roses Are Red, Violets Are Blue... But Should VQA Expect Them to?}'' in \emph{CVPR}, 2021, pp. 2776--2785.

\bibitem{anderson2018bottom}
P.~Anderson, X.~He, C.~Buehler, D.~Teney, M.~Johnson, S.~Gould, and L.~Zhang, ``{Bottom-up and Top-down Attention for Image Captioning and Visual Question Qnswering},'' in \emph{CVPR}, 2018, pp. 6077--6086.

\bibitem{jiang2020defense}
H.~Jiang, I.~Misra, M.~Rohrbach, E.~Learned-Miller, and X.~Chen, ``{In Defense of Grid Features for Visual Question Answering},'' in \emph{CVPR}, 2020, pp. 10\,267--10\,276.

\bibitem{garderes2020conceptbert}
F.~Gard{\`e}res, M.~Ziaeefard, B.~Abeloos, and F.~Lecue, ``{Conceptbert: Concept-aware Representation for Visual Question Answering},'' in \emph{Findings of the EMNLP}, 2020, pp. 489--498.

\bibitem{guo2021re}
W.~Guo, Y.~Zhang, J.~Yang, and X.~Yuan, ``{Re-attention for Visual Question Answering},'' \emph{IEEE TIP}, vol.~30, pp. 6730--6743, 2021.

\bibitem{wu2017visual}
Q.~Wu, D.~Teney, P.~Wang, C.~Shen, A.~Dick, and A.~Van Den~Hengel, ``{Visual Question Answering: A Survey of Methods and Datasets},'' \emph{CVIU}, vol. 163, pp. 21--40, 2017.

\bibitem{noh2016image}
H.~Noh, P.~H. Seo, and B.~Han, ``{Image Question Answering Using Convolutional Neural Network with Dynamic Parameter Prediction},'' in \emph{CVPR}, 2016, pp. 30--38.

\bibitem{gao2015you}
H.~Gao, J.~Mao, J.~Zhou, Z.~Huang, L.~Wang, and W.~Xu, ``{Are You Talking to A Machine? Dataset and Methods for Multilingual Image Question Answering},'' in \emph{NeurIPS}, 2015, pp. 2296--2304.

\bibitem{9466370}
J.~Ma, J.~Liu, Y.~Wang, J.~Li, and T.~Liu, ``{Relation-Aware Fine-Grained Reasoning Network for Textbook Question Answering},'' \emph{TNNLS}, vol.~34, no.~1, pp. 15--27, 2023.

\bibitem{chen2022grounding}
C.~Chen, S.~Anjum, and D.~Gurari, ``{Grounding Answers for Visual Questions Asked by Visually Impaired People},'' in \emph{CVPR}, 2022, pp. 19\,098--19\,107.

\bibitem{andreas2016learning}
J.~Andreas, M.~Rohrbach, T.~Darrell, and D.~Klein, ``{Learning to Compose Neural Networks for Question Answering},'' in \emph{NAACL}, 2016, pp. 1545--1554.

\bibitem{chen2020counterfactual}
L.~Chen, X.~Yan, J.~Xiao, H.~Zhang, S.~Pu, and Y.~Zhuang, ``{Counterfactual Samples Synthesizing for Robust Visual Question Answering},'' in \emph{CVPR}, 2020, pp. 10\,800--10\,809.

\bibitem{li2022invariant}
Y.~Li, X.~Wang, J.~Xiao, W.~Ji, and T.-S. Chua, ``{Invariant Grounding for Video Question Answering},'' in \emph{CVPR}, 2022, pp. 2928--2937.

\bibitem{niu2021introspective}
Y.~Niu and H.~Zhang, ``{Introspective Distillation for Robust Question Answering},'' in \emph{NeurIPS}, 2021, pp. 16\,292--16\,304.

\bibitem{Kolling}
C.~Kolling, M.~More, N.~Gavenski, E.~Pooch, O.~Parraga, and R.~C. Barros, ``{Efficient Counterfactual Debiasing for Visual Question Answering},'' in \emph{WACV}, 2022, pp. 3001--3010.

\bibitem{chen2022rethinking}
L.~Chen, Y.~Zheng, and J.~Xiao, ``{Rethinking Data Augmentation for Robust Visual Question Answering},'' in \emph{Computer Vision--ECCV 2022: 17th European Conference, Tel Aviv, Israel, October 23--27, 2022, Proceedings, Part XXXVI}, 2022, pp. 95--112.

\bibitem{NEURIPS20200}
D.~Teney, E.~Abbasnejad, K.~Kafle, R.~Shrestha, C.~Kanan, and A.~van~den Hengel, ``{On the Value of Out-of-distribution Testing: an Example of Goodhart\textquotesingle's Law},'' in \emph{NeurIPS}, 2020, pp. 407--417.

\bibitem{liang2020learning}
Z.~Liang, W.~Jiang, H.~Hu, and J.~Zhu, ``{Learning to Contrast the Counterfactual Samples for Robust Visual Question Answering},'' in \emph{EMNLP}, 2020, pp. 3285--3292.

\bibitem{gupta2022swapmix}
V.~Gupta, Z.~Li, A.~Kortylewski, C.~Zhang, Y.~Li, and A.~Yuille, ``{SwapMix: Diagnosing and Regularizing the Over-reliance on Visual Context in Visual Question Answering},'' in \emph{CVPR}, 2022, pp. 5078--5088.

\bibitem{si2022towards}
Q.~Si, Y.~Liu, F.~Meng, Z.~Lin, P.~Fu, Y.~Cao, W.~Wang, and J.~Zhou, ``{Towards Robust Visual Question Answering: Making the Most of Biased Samples via Contrastive Learning},'' \emph{arXiv preprint arXiv:2210.04563}, 2022.

\bibitem{lao2022vqa}
M.~Lao, Y.~Guo, W.~Chen, N.~Pu, and M.~S. Lew, ``{VQA-BC: Robust Visual Question Answering Via Bidirectional Chaining},'' in \emph{ICASSP}, 2022, pp. 4833--4837.

\bibitem{xiong2018dcn}
C.~Xiong, V.~Zhong, and R.~Socher, ``{DCN+: Mixed Objective and Deep Residual Coattention for Question Answering},'' in \emph{ICLR}, 2018.

\bibitem{yu2018qanet}
A.~W. Yu, D.~Dohan, M.-T. Luong, R.~Zhao, K.~Chen, M.~Norouzi, and Q.~V. Le, ``{Qanet: Combining Local Convolution with Global Self-attention for Reading Comprehension},'' in \emph{ICLR}, 2018.

\bibitem{chen2021bidirectional}
S.~Chen, Y.~Wang, J.~Liu, and Y.~Wang, ``{Bidirectional Machine Reading Comprehension for Aspect Sentiment Triplet Extraction},'' in \emph{AAAI}, 2021, pp. 12\,666--12\,674.

\bibitem{li2021asynchronous}
R.~Li, L.~Wang, S.~Wang, and Z.~Jiang, ``{Asynchronous Multi-grained Graph Network For Interpretable Multi-hop Reading Comprehension},'' in \emph{IJCAI}, 2021, pp. 3857--3863.

\bibitem{3052594}
Y.~Liu, Y.~Liu, K.~Zhou, M.~Zhang, and S.~Ma, ``{Detecting Collusive Spamming Activities in Community Question Answering},'' in \emph{www}, 2017, pp. 1073--1082.

\bibitem{wang2017machine}
S.~Wang and J.~Jiang, ``{Machine Comprehension Using Match-LSTM and Answer Pointer},'' in \emph{ICLR}, 2017, pp. 1--15.

\bibitem{vinyals2015pointer}
O.~Vinyals, M.~Fortunato, and N.~Jaitly, ``{Pointer Networks},'' in \emph{NeurIPS}, 2015, pp. 2692--2700.

\bibitem{wang2017gated}
W.~Wang, N.~Yang, F.~Wei, B.~Chang, and M.~Zhou, ``{Gated Self-matching Networks for Reading Comprehension and Question Answering},'' in \emph{ACL}, 2017, pp. 189--198.

\bibitem{SeoKFH17}
M.~J. Seo, A.~Kembhavi, A.~Farhadi, and H.~Hajishirzi, ``{Bidirectional Attention Flow for Machine Comprehension},'' in \emph{ICLR}, 2017.

\bibitem{hu2018reinforced}
M.~Hu, Y.~Peng, Z.~Huang, X.~Qiu, F.~Wei, and M.~Zhou, ``{Reinforced Mnemonic Reader for Machine Reading Comprehension},'' in \emph{IJCAI}, 2018, pp. 4099--4106.

\bibitem{yuqanet}
A.~W. Yu, D.~Dohan, M.-T. Luong, R.~Zhao, K.~Chen, M.~Norouzi, and Q.~V. Le, ``{QANet: Combining Local Convolution with Global Self-Attention for Reading Comprehension},'' in \emph{ICLR}, 2018.

\bibitem{chen2021adaptive}
N.~Chen, F.~Liu, C.~You, P.~Zhou, and Y.~Zou, ``{Adaptive bi-directional attention: Exploring multi-granularity representations for machine reading comprehension},'' in \emph{ICASSP}, 2021, pp. 7833--7837.

\bibitem{shinoda2022look}
K.~Shinoda, S.~Sugawara, and A.~Aizawa, ``{Look to the Right: Mitigating Relative Position Bias in Extractive Question Answering},'' \emph{arXiv preprint arXiv:2210.14541}, 2022.

\bibitem{HastieTF09}
T.~Hastie, R.~Tibshirani, and J.~H. Friedman, \emph{{The Elements of Statistical Learning: Data Mining, Inference, and Prediction, 2nd Edition}}.\hskip 1em plus 0.5em minus 0.4em\relax Springer, 2009.

\bibitem{lin2017focal}
T.-Y. Lin, P.~Goyal, R.~Girshick, K.~He, and P.~Doll{\'a}r, ``Focal loss for dense object detection,'' in \emph{ICCV}, 2017, pp. 2980--2988.

\bibitem{garipov2018loss}
T.~Garipov, P.~Izmailov, D.~Podoprikhin, D.~P. Vetrov, and A.~G. Wilson, ``{Loss Surfaces, Mode Connectivity, and Fast Ensembling of DNNs},'' in \emph{NeurIPS}, 2018, pp. 8803--8812.

\bibitem{Chen18g}
D.~Chen, ``{Neural Reading Comprehension and Beyond},'' Ph.D. dissertation, Stanford University, 2018.

\bibitem{he2019unlearn}
H.~He, S.~Zha, and H.~Wang, ``{Unlearn Dataset Bias in Natural Language Inference by Fitting the Residual},'' in \emph{EMNLP-IJCNLP}, 2019, pp. 132--142.

\bibitem{yang2022vision}
J.~Yang, J.~Duan, S.~Tran, Y.~Xu, S.~Chanda, L.~Chen, B.~Zeng, T.~Chilimbi, and J.~Huang, ``{Vision-language Pre-training with Triple Contrastive Learning},'' in \emph{CVPR}, 2022, pp. 15\,671--15\,680.

\bibitem{gao2019dynamic}
P.~Gao, Z.~Jiang, H.~You, P.~Lu, S.~C. Hoi, X.~Wang, and H.~Li, ``Dynamic fusion with intra-and inter-modality attention flow for visual question answering,'' in \emph{CVPR}, 2019, pp. 6639--6648.

\bibitem{wang2023learning}
T.-J.~J. Wang, J.~Laaksonen, T.~Langer, H.~Arponen, and T.~E. Bishop, ``{Learning by Hallucinating: Vision-Language Pre-training with Weak Supervision},'' in \emph{WACV}, 2023, pp. 1073--1083.

\bibitem{yang2019xlnet}
Z.~Yang, Z.~Dai, Y.~Yang, J.~Carbonell, R.~R. Salakhutdinov, and Q.~V. Le, ``{XLNet: Generalized Autoregressive Pretraining for Language Understanding},'' in \emph{NeurIPS}, 2019, pp. 5753--5763.

\end{thebibliography}

\end{document}